\documentclass[11pt]{article}
\usepackage{floatrow}
\newfloatcommand{capbtabbox}{table}[][\FBwidth]
\usepackage[utf8]{inputenc}

\usepackage{wrapfig}
\usepackage{amsfonts}
\usepackage{amsmath}
\usepackage{amsthm}
\usepackage{color}
\usepackage[bottom]{footmisc}
\usepackage{newtxtext}
\usepackage{subfig}
\usepackage{graphicx}
\usepackage{algorithm}
\usepackage{algorithmic}
\usepackage{placeins}
\usepackage{booktabs}
\usepackage{floatrow}
\usepackage{blindtext}
\usepackage{etoolbox}  % patch def of algorithmic environment

\makeatletter
\patchcmd{\algorithmic}{\addtolength{\ALC@tlm}{\leftmargin} }{\addtolength{\ALC@tlm}{\leftmargin}}{}{}
\makeatother
\usepackage{array}
\usepackage{eqparbox}

\usepackage[margin=1in]{geometry}
\usepackage{mathtools}
\usepackage{booktabs}
\usepackage{authblk}
\usepackage{url}
\graphicspath{{figs/}}

\newcommand{\gemnet}{\textsc{Gemtelligence}}
\definecolor{blueoff}{RGB}{89,117,164}
\definecolor{brownoff}{RGB}{204,137,99}
\definecolor{greenoff}{RGB}{95,158,110}
\definecolor{redoff}{RGB}{181,93,96}
\definecolor{FTIR_col}{RGB}{147,120,96}
\definecolor{ED_col}{RGB}{196,79,83}
\definecolor{UV_col}{RGB}{129,114,179}
\definecolor{ALL_but_icp_col}{RGB}{85,168,104}
\definecolor{ICP_col}{RGB}{221,132,82}
\definecolor{ALL_col}{RGB}{76,114,176}

\usepackage{graphicx}
\usepackage{booktabs} % for professional tables
\usepackage{multirow}
\usepackage{csquotes}
\usepackage{listings}
\usepackage{array}
\usepackage{tabularx}
\usepackage{multirow}
\usepackage{multicol}
\usepackage{hhline}
\usepackage{caption}
\usepackage{comment}
\usepackage{float}
\usepackage{lineno}
\usepackage{hyperref}
\usepackage{floatrow}

\title{\gemnet: Accelerating Gemstone classification with Deep Learning}
\author[3]{Tommaso Bendinelli$^*$}
\author[1,3]{Luca Biggio$^*$}
\author[2]{Daniel Nyfeler}
\author[2]{Abhigyan Ghosh}
\author[2]{Peter Tollan}
\author[3]{Moritz Alexander Kirschmann}
\author[4]{Olga Fink}

\affil[1]{Data Analytics Lab, ETH, Zürich, Switzerland}
\affil[2]{Gübelin Gem Lab Ltd, Lucerne, Switzerland}
\affil[3]{CSEM SA, Alpnach, Switzerland}
\affil[4]{EPFL, Lausanne, Switzerland}
\date{}

\begin{document}
\maketitle

\begin{abstract}
The value of luxury goods, particularly investment-grade gemstones, is greatly influenced by their origin and authenticity, sometimes resulting in differences worth millions of dollars. Traditionally, human experts have determined the origin and detected treatments on gemstones through visual inspections and a range of analytical methods. However, the interpretation of the data can be subjective and time-consuming, resulting in inconsistencies. In this study, we propose \gemnet, a novel approach based on deep learning that enables accurate and consistent origin determination and treatment detection. \gemnet{} comprises convolutional and attention-based neural networks that process heterogeneous data types collected by multiple instruments. Notably, the algorithm demonstrated comparable predictive performance to expensive laser-ablation inductively-coupled-plasma mass-spectrometry (ICP-MS) analysis and visual examination by human experts, despite using input data from relatively inexpensive analytical methods. Our innovative methodology represents a major breakthrough in the field of gemstone analysis by significantly improving the automation and robustness of the entire analytical process pipeline.
\end{abstract}

\section*{Introduction}
Gemstones, both natural and synthetic, are highly prized for their rarity and beauty and are commonly used in jewelry for both adornment and investment purposes. Some of these minerals can be worth over a million dollars per gram, making them some of the most concentrated physical capital in the world. In addition to factors such as species and aesthetic quality, the value of a gemstone is also influenced by its geographic origin and any potential treatments it may have undergone after being mined. 
Identifying these treatments, which can include exposure to electromagnetic radiation \cite{phichaikamjornwut2019conclusive}, heating \cite{jaliya2020characterization,emmett2003beryllium}, or the infusion of oils or other substances \cite{johnson1999identification}, is crucial for determining the true value of a precious stone and upholding consumer trust in the jewelry industry. Unfortunately, the artisanal nature of gemstone mining results in a fragmented and opaque supply chain, making it difficult to reliably track the origin and treatment of individual stones. \\
To minimize investment risks, buyers and sellers  often require that gemstones are accompanied by an independent laboratory report that confirms the physical characteristics, treatment status, and geographic source of the stones.
In the conventional practice of determining the authenticity and source of gemstones, skilled human experts with two to six years of training in gemology play a pivotal role \cite{pardieu2020field}. These specialists have traditionally relied on optical microscopy to identify structures and inclusions within the gemstones that can provide clues about their origin and any treatments they may have undergone. However, this task is exceptionally difficult, as gemstones from different locations may exhibit strikingly similar features due to shared geological histories \cite{dissanayake1999sri}. Furthermore, with the advancement of treatment techniques for gemstones, the detection of such treatments has become increasingly challenging for even the most experienced human examiners \cite{rankin2006gubelin, hughes2019madagascar}. \\

\noindent
In response to these pressing challenges and the demands of the industry, state-of-the-art gemology laboratories have introduced a variety of new analytical instruments in their regular workflow \cite{groat2019review}, including ultraviolet-visible-near-infrared spectroscopy (UV) \cite{scarani2017gemological}, Fourier-transform infrared-spectroscopy (FTIR) \cite{kitawaki2006identification}, energy-dispersive X-ray fluorescence (XRF) \cite{joseph2000characterization} and laser-ablation inductively-coupled-plasma mass spectrometry (ICP-MS) \cite{guillong2001quasi}. The UV, FTIR, and XRF devices come at a combined cost of around 200,000 USD and can be operated by technical staff with just a brief introductory training. However, the ICP-MS instrument alone, albeit providing the most comprehensive data for identifying origins, costs approximately 500,000 USD and demands one or more qualified operators with extensive theoretical and practical training. In addition, ICP-MS uses a laser to ablate a small volume from the gems, making it, therefore, destructive on a micro-scale level \cite{rossman2009geochemistry}. 
Despite all these data sources, the tasks of determining the origin and detecting treatment with high accuracy remain very challenging because the differences in physical, spectroscopic, and chemical properties between stones from different origins are often subtle. 
This is especially true for blue sapphires, which are among the "big 3" gemstone species that top-tier gemology laboratories most often assess \cite{palke2019geographic}. Moreover, even with strict adherence to laboratory protocols, such as restricting experts' access to analytical results during microscopic examinations and ensuring at least two independent conclusions, it remains a challenging task to obtain a consistent outcome from a combination of visual and analytical data. Therefore, the final decision is inevitably susceptible to subjective biases. In addition, as gemstones are long-term investments, they re-enter the market from time to time and new evaluations of the same stones are conducted. Inconsistencies in the origin or treatment determination of the same stone over time, which massively affect a stone's value, can undermine the confidence in this asset class. Crucially, some gemstones cannot be classified unambiguously and with high confidence by the experts. While ICP-MS analysis can help mitigate this issue, it can also result in significant costs. For all these reasons, the advancement of innovative methods that effectively leverage data obtained from affordable instruments while maximizing accuracy and robustness is of considerable practical significance. \\

\noindent
Modern machine- and deep-learning algorithms have revolutionized the analysis and interpretation of large and complex datasets in various fields, including but not limited to material science \cite{matsci1,matsci2,matsci3}, geoscience \cite{clisci1,clisci2}, and computational chemistry \cite{dg1,dg2,dg3} allowing for more accurate and efficient data processing. However, their application to gemology is still in its infancy. Conventional machine-learning techniques in gemology that typically involve feature extraction methods followed by simple downstream algorithms, have provided promising results in automating various gemological tasks, such as categorizing gemstones by type and shape \cite{chow2021automatic,botejuresearch,ostreika2021classification}, distinguishing real and synthetic gemstones \cite{qiu2019feasibility}, and even more complicated tasks like grading gemstones \cite{wang2015automated}, \cite{amarasekara2021convolutional}. 
Nevertheless, such techniques are restricted to analyzing only one type of data source at a time, such as images, spectra, or tabular data \cite{shukla2022artificial}, limiting their capability of detecting artificial treatments or correctly identifying the origin of the gemstone. As such, these challenging tasks still heavily rely on human expertise.  \\

\begin{figure}[H]
    \centering
    \includegraphics[scale = 0.69]{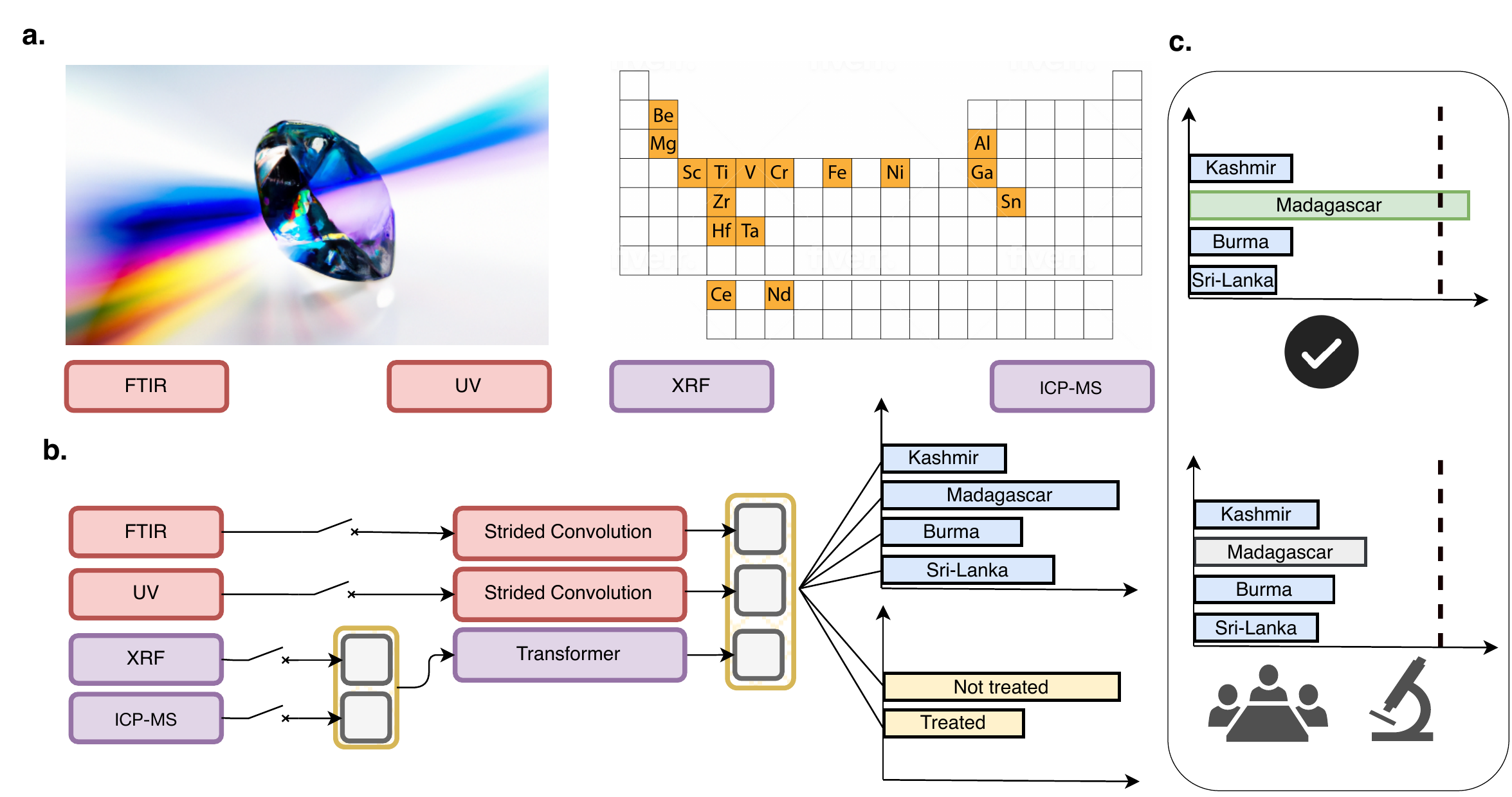}    
    \caption{\textbf{a.} \gemnet{} can process measurements from four distinct data sources: FTIR and UV (spectroscopy analysis) and ICP-MS and XRF (elemental analysis). \textbf{b.} With a negligible inference time compared to human experts, it can predict the probability of a gemstone's origin or whether it has undergone heat treatment. Not all data types are required for inference; missing sources can be masked out as illustrated by switch symbols in the figure. \textbf{c.} If the maximum probability exceeds a predefined threshold (top panel), the stone prediction can be confidently accepted. If the maximum probability falls below the threshold (bottom panel), however, the output should be discarded and the stone should be further analyzed via standard methods such as microscopy and expert analysis. The value of the threshold, selected during the confidence-thresholding phase, determines the balance between the number of stones that can be processed automatically and the accuracy achieved by the model. }
    \label{fig:main}
\end{figure}

\noindent
Herein, we propose \gemnet{} (Fig. \ref{fig:main}), a deep learning-based method that automates the determination of the country of origin (OD) and detection of treatment (TD)  of gemstones at a fraction of the time and cost, outperforming human experts' evaluations and without relying on costly measurements. This study is the first of its kind to address both OD and TD of valuable gemstones using a novel deep learning approach specifically tailored to handle varied and multi-modal analytical data acquired from different testing devices.
Crucially, we conduct our experimental evaluation on a large collection of high-quality gemstones, thereby, allowing us to examine the performance of our algorithm in real-world scenarios. A part of the aforementioned data, along with the source code of our model, will be made available for public use to facilitate the benchmarking and reproducibility of the results presented in this work.\\

\noindent
The primary innovation of the proposed approach lies in its multi-modal design, which is custom-tailored to effectively process and integrate varied and diverse analytical data acquired from different instruments.
{\gemnet} consists of a combination of strided convolutional neural networks \cite{dumoulin2016guide} and a variant of the popular Transformer architecture \cite{vaswani2017attention}. Particularly, for processing spectral data, we draw inspiration from the work of Ho et al. \cite{ho2019rapid} and use a modified version of their architecture with a bigger kernel size to increase the receptive field in order to capture more global features. The transformer-like component of \gemnet{} is used to process tabular data and is based on the architecture introduced in \cite{somepalli2021saint}. The final architecture combines all these elements in a single model enabling end-to-end multi-modal training.  As robustness and consistency are key desiderata in gemstone analysis, we include an additional confidence-thresholding scheme in our pipeline allowing users to control the trade-off between the degree of automation provided by \gemnet{} (i.e. the number of stones that can be processed automatically) and its level of accuracy. As illustrated in Fig. \ref{fig:main} (c), given a new test stone, we compare the most likely prediction of \gemnet{} with a threshold value and we accept the prediction only if the corresponding probability exceeds this threshold. The value of this threshold is determined by imposing that the trained model attains the desired accuracy level on the training set. More details about the model architecture and the confidence-thresholding procedure can be found in the Methods section. \\

\noindent
Our work contributes to the burgeoning field of laboratory automation \cite{pyzer2022accelerating,vaucher2020automated,vaucher2021inferring,naugler2019automation}, which has seen a rising focus on leveraging Artificial Intelligence (AI) techniques to streamline the time-consuming and repetitive pipelines typical of applied scientific research. As shown by our empirical evaluation, the implementation of {\gemnet} in gemological laboratories can assist human experts in the time-consuming task of data assessment and interpretation, allowing them to focus on more high-value activities, including research and development. \\

\section*{Results}
In this section, we present the results of \gemnet{} for the two challenging tasks of OD and TD of blue sapphires. Blue sapphire is a type of corundum, $\mathrm{Al_2O_3}$, with a blue hue caused by the presence of trace amounts of Fe and Ti. Our focus on sapphires stems from two key factors. Firstly, sapphires are widely acknowledged to present more difficulties in achieving accurate OD than other gemstones \cite{palke2019geographic}. Secondly, the TD of sapphires has not been researched as extensively as that of other gemstones like rubies.
In the following, we first introduce the tasks, datasets, training, and testing pipelines. Then, we assess \gemnet{}'s performance in processing diverse multi-modal data, through various ablation studies and in comparison to human experts.

\subsection*{Background and Experimental setup}
\paragraph{Origin Determination.} The problem of OD can be cast as a classification task: based on the laboratory tests performed on a particular gemstone, the goal is to determine its geographical origin out of a discrete set of candidates. This study focuses on the top four significant sources of high-value blue sapphires, which make up over 90\% of the market's high-quality blue sapphire volume. These sapphires were created through metamorphism and are sourced from Kashmir\footnote{In the present context, the term Kashmir refers to the Jammu and Kashmir region at the northern tip of India, between Pakistan and China. There, in the Padar area, in the Kudi valley near the village of Sumjam, high-quality blue sapphires were found during a few years only in the second half of the 19$^{th}$ century.}, Burma/Myanmar, Sri Lanka, and Madagascar. The geological environment and rock varieties in which blue sapphires develop have specific attributes \cite{giuliani2019geology}, resulting in slightly varying gemological characteristics \cite{palke2019geographic}. This allows for distinguishing the origins by examining these features. However, identifying the origin (OD) remains a challenging task since the physical, chemical, and spectroscopic properties of blue sapphires from distinct sources often have considerable overlap \cite{groat2019review}. Furthermore, blue sapphire has a very limited range of routinely detectable trace elements in comparison to other gemstones, substantially limiting the effectiveness of traditional multi-component statistical analysis for OD. 
Microscopy and ICP-MS are widely considered the most reliable analytical evaluations for OD. UV and XRF are also employed, though their analysis results are not as reliable as those from microscopy and ICP-MS. The usefulness of FTIR in identifying the origin of sapphires from metamorphic rocks is limited as those sapphires have very few origin-specific phases that can be detected by FTIR, such as aluminum oxides-hydroxides and structurally bonded OH groups \cite{smith1995contribution}.
\paragraph{Artificial Heat Treatment Detection.} Artificial heat treatment is the process of heating gemstones to improve their visual appearance, clarity, and color. To determine if a stone has been artificially heat treated, the primary method used is visual microscopic inspection complemented with spectral measurement techniques. Heat treatment may result in structural changes, particularly in microscopic or submicroscopic small inclusions. The term \enquote{inclusions} refers to phases (solid, liquid, or gaseous) that are trapped or formed in the crystal during or after the growth of the stone in the earth \cite{gubelin1953inclusions}. For instance, an inclusion in a gemstone might become unstable when heated and begins to disintegrate. FTIR and UV analysis are typically employed to support heat treatment detection on sapphires. Since elemental analysis methods such as XRF and ICP-MS cannot capture the underlying physical change of a stone undergoing heat treatment, they are not used for this task. Hence, we also avoid using them as input to our algorithm in order to prevent the introduction of spurious correlations. We frame the TD problem as a binary classification task, comprising a \enquote{treated} and a \enquote{non-treated} class.

\paragraph{Training and Testing Datasets.} 
The data used for training \gemnet{} comprises over 5500 blue sapphire records obtained from the Gubelin Gem Lab over the course of ten years using various protocols differing from each other in the number and type of analyses performed. A full protocol involves taking two optical and two chemical spectroscopic measurements (UV, FTIR, XRF, and ICP-MS) for each gemstone. However, in cases where a reduced protocol is followed, certain analyses may be omitted. To assign OD and TD labels, two or more experienced professionals initially assign candidates for each measurement independently and then visually examine the stone. By comparing the outcomes of each independent analysis, a consensus is reached for the final assignment, which is regarded as our ground truth. We evaluate the performance of both OD and TD models using five-fold cross-validation. The main advantage of this procedure is that it provides a more reliable estimate of the model's performance than a simple train/test split, especially when the dataset is relatively small, as in our case. The data are randomly split into five sets, of which one is used for testing and the remaining four are used for training. This process is repeated for each fold. The training data are used to train, validate, and calibrate the model, while the test data are employed to measure the model's performance. In order to ensure a rigorous analysis, stones from the test set that do not meet all the following criteria are excluded: 1) each measurement is examined independently from the others; 2) all possible relevant measurements are taken; 3) two expert gemologists independently reach the same conclusion via visual inspection in TD and the results obtained from ICP-MS and visual inspection match in OD.
Additional details can be found in Supplementary Note 5.

\subsection*{Performance Evaluation}

\begin{figure}[H]
    \centering
    \includegraphics[width=0.45\textwidth,height=6cm]{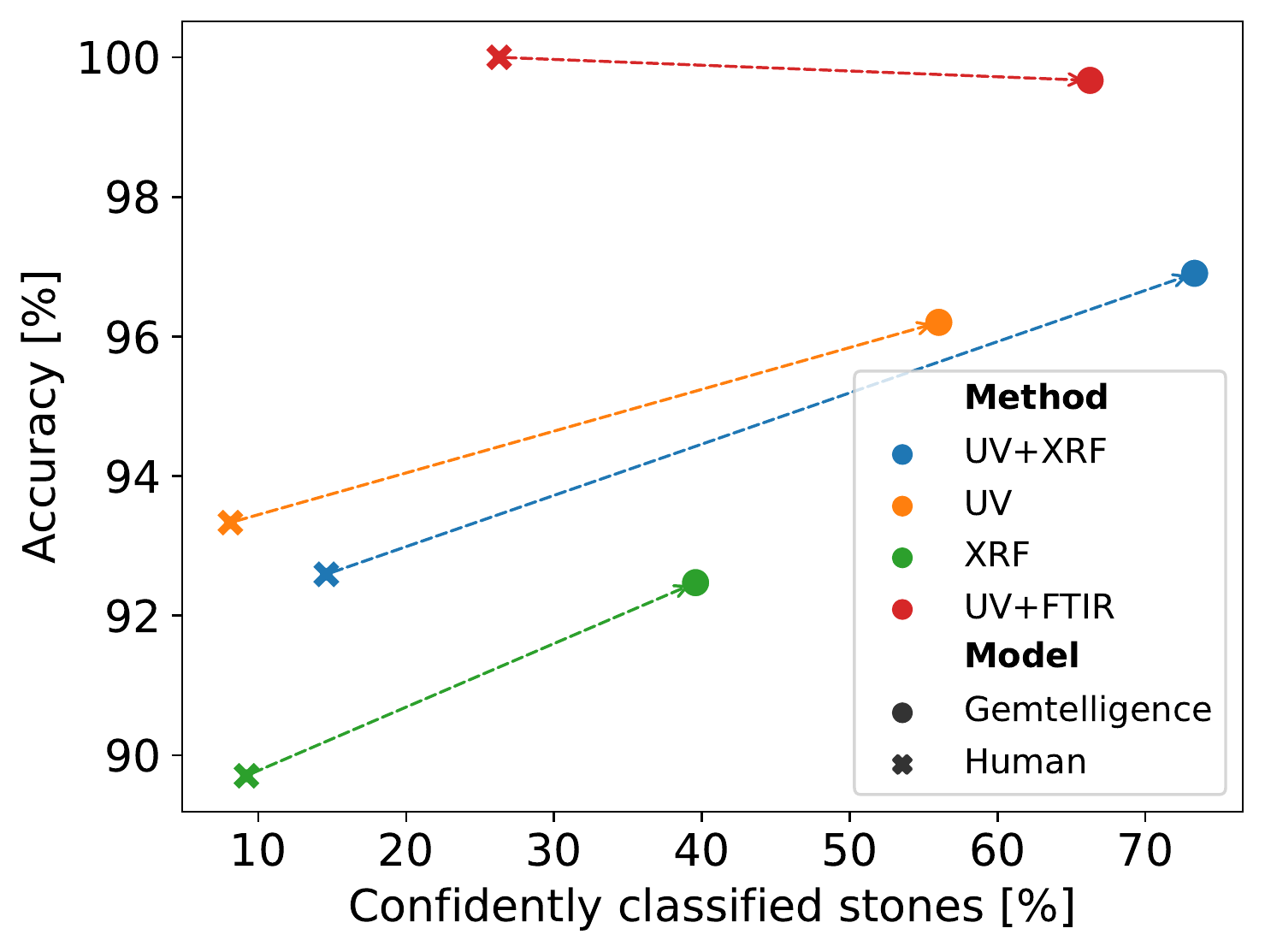}
    \caption{Comparison between human experts (represented by crosses) and \gemnet{} (represented by circles) in terms of the size of the subset of stones that have been confidently classified (on the x-axis) and the corresponding level of accuracy achieved for this subset (on the y-axis). Each color corresponds to a different combination of data sources. All the combinations apart from the red one (UV+FTIR) are used for OD while UV+FTIR is used for TD. The dashed lines are used to highlight the performance change between humans and our model. The results in the plot are obtained by evaluating the performance of experts and \gemnet{} on test data.} 
\label{fig:bar_situation}
\end{figure}

\paragraph{AI-supported decision system.}
To begin our analysis, we compare the performance of \gemnet{} with that of human gemologists on various combinations of data sources and tasks. The gemologists follow a strict procedure throughout the process of analyzing the gemstones. First, each data source (e.g. XRF) is observed independently and without access to other information. Second, based solely on this data source, a preliminary conclusion is made regarding the gemstone's origin and any potential heat treatment it may have undergone. We use these sub-conclusions to create statistics and perform comparisons between \gemnet{} and human experts using individual and combined data sources. \\
Fig. \ref{fig:bar_situation} shows a comparison between \gemnet{} and human experts on the OD and TD tasks in terms of the number of stones they can confidently classify based on single or combined data sources
\footnote{For the sake of fairness, we conduct this comparison by considering only the data sources that were not used to determine the label of the test data. Specifically, we do not use ICP-MS data for OD as the test set was created such that the final conclusion had to match the ICP-MS sub-conclusion, which is regarded as the most reliable data source. For TD, the gemologists examine UV and FTIR spectra simultaneously, so we do not have access to separate statistics.}
and the obtained levels of accuracy.  

\textbf{}

\noindent
For \gemnet{}, we refer to a stone being confidently classified if the probability of the model associated with its final prediction exceeds the threshold value (see Fig. \ref{fig:main} c.). In this specific experiment, the threshold has been determined by calibrating the model on the training data to match or surpass the accuracy levels reached by human experts on the test data. For OD, human experts confidently classify a stone if they return a single prediction, rather than a list of possible candidates. In the case of TD, the expert is not confident if the uncertainty is too high to draw a final conclusion. \\
For all the considered combinations of data sources, \gemnet{} can provide confident predictions on substantially larger sets of stones (x-axis) than human experts, who are often  unable to draw definitive conclusions due to uncertainty. Remarkably, \gemnet{} also achieves either comparable or significantly higher accuracy levels (y-axis) than human experts while delivering a final conclusion on much larger groups of stones. This experiment demonstrates that \gemnet{} improves the level of automation in gemstone analysis, significantly reducing the analysis time compared to human experts (who typically take several hours per stone) while achieving comparable or even higher levels of accuracy.\\
We proceed with our analysis by exploring how different threshold values yield different trade-offs between accuracy and automation. A higher threshold value results in more stones requiring further human expert analysis but potentially higher accuracy, while a lower threshold value may require less human expert analysis but potentially lower accuracy. 
Table \ref{table1:results} presents the performance of our model, in three operating setups, namely \texttt{None}, \texttt{Mode 1} and \texttt{Mode 2}, each offering different trade-offs between automation and accuracy. For the last two modes, the threshold of \gemnet{} is chosen to ensure a specific accuracy rate (98\% for \texttt{Mode 1} and 99\% for \texttt{Mode 2}) on the training stones that have a classification probability above this threshold. For more details on the confidence-thresholding procedure, please refer to Methods section.
For the \texttt{None} setup (second and fifth columns in Table \ref{table1:results}), the model is not calibrated since the threshold is set to zero. This means that all predictions are accepted, resulting in the complete automation of the process (100\% stones above the threshold) at the cost of higher error rates. However, in scenarios where the model's prediction uncertainty is higher, this configuration lacks preventive measures to mitigate the risk of potential errors.
For the other two setups, the threshold is non-zero, resulting in fewer accepted predictions and increased accuracy compared to the \texttt{None} setup. The lower threshold for \texttt{Mode 1} compared to \texttt{Mode 2} results in a significant increase in the number of confidently classified stones for both tasks. Compared to \texttt{Mode 2}, this setup greatly reduces the workload of gemologists, as the inference time of \gemnet{} is negligible (less than a second), whereas taking a final decision on a single stone can take several hours for human experts. Nevertheless, \texttt{Mode 1} also leads to a slight reduction in test accuracy, although the results remain favorable and comparable to human capabilities. \\
In the field of gemstone analysis, the level of automation and accuracy of predictions can significantly impact the workload of gemologists and the value of stones. Depending on the specific use case, it may be advantageous to prioritize one over the other. As incorrect evaluations can significantly impact stone prices, \texttt{Mode 2} mode represents a more conservative and low-risk configuration as it results in high accuracy levels, despite decreasing the level of automation of the model (number of stones confidently classified).

\begingroup
\setlength{\tabcolsep}{1pt}
\renewcommand{\arraystretch}{1}
\begin{table}[H]
\centering
\makebox[0.5\textwidth][c]{%
\begin{tabularx}{\textwidth}{@{}l|XXX|XXX@{}}
                   & \multicolumn{3}{c}{\textbf{Origin Determination}}& \multicolumn{3}{c}{\textbf{Heat Treatment Detection}} \\ \hline\hline
\textbf{\gemnet{} setup} & \hspace{10pt}\small{\texttt{None}} & \footnotesize{\texttt{Mode 1}} & \footnotesize{\texttt{Mode 2}} & \hspace{10pt}\small{\texttt{None}} & \footnotesize{\texttt{Mode 1}} & \footnotesize{\texttt{Mode 2}} \\ \hline\hline
\textbf{Calibration accuracy}  &\hspace{10pt} None & 98\% & 99\% &\hspace{10pt} None & 98\% & 99\% \\
\textbf{Stones above threshold}  &\hspace{8pt} 100.0\% & 74.2\% &38.5\% &\hspace{10pt} 100.0\%  & 97.4\% & 95.5\% \\
\textbf{Test accuracy} & \hspace{10pt}90.69\% & 96.8\% & 99.1\% &\hspace{10pt} 98.03\% & 98.7\% & 98.9\% \\
\end{tabularx}
\caption{Calibration accuracy used to determine the threshold (first row), the corresponding number of analyzed stones (second row), and test accuracy (third row) for the three considered operating setups both for OD and TD. OD was performed using UV and XRF and TD was done using UV and FTIR.
}
\label{table1:results}
}
\end{table}
\endgroup
\noindent

\begin{figure}[H]
    \centering
    
    \includegraphics[scale = 0.43]{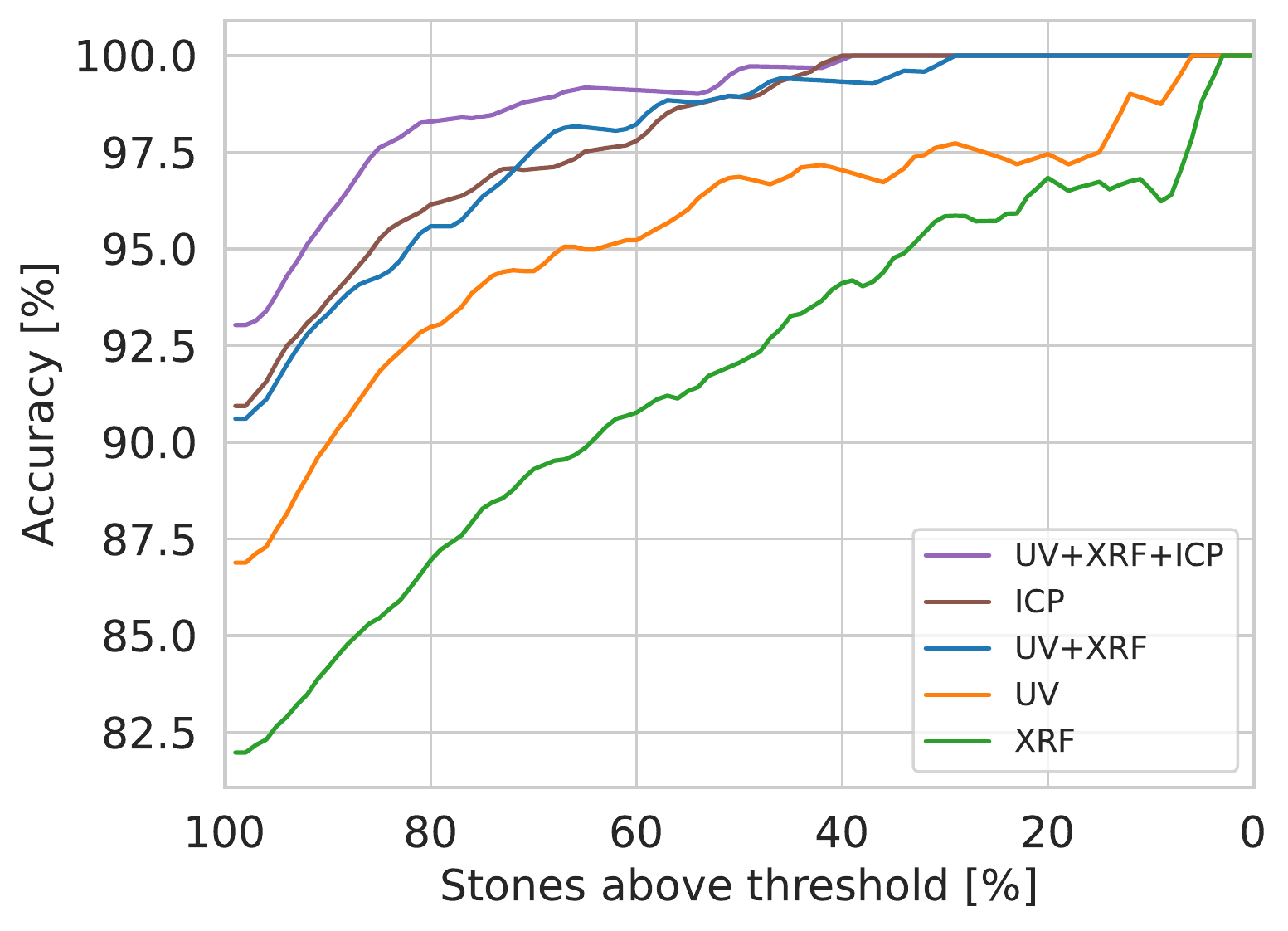} 
    \includegraphics[scale = 0.43]{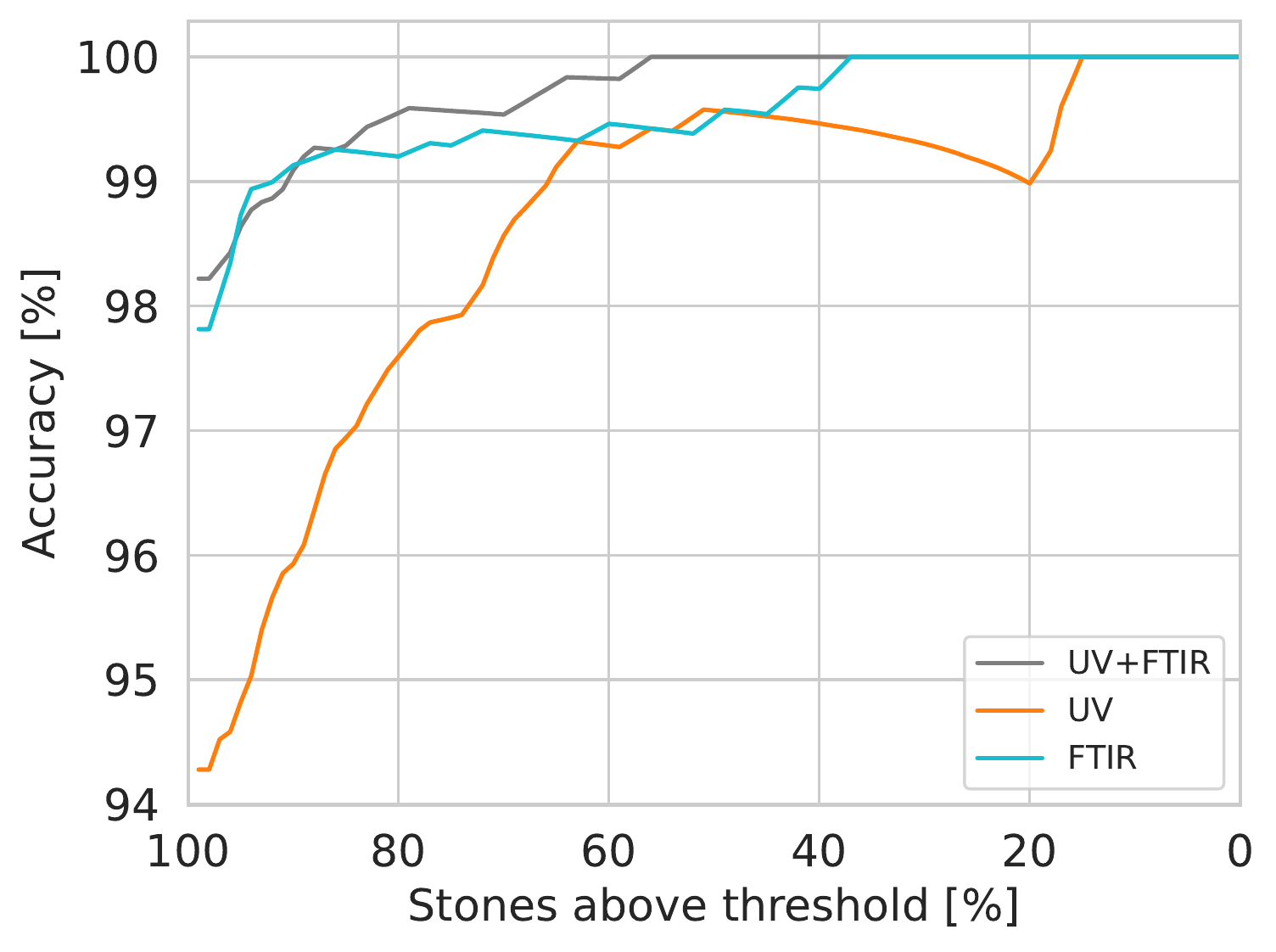} 
    
    \caption{Accuracy (\%) vs. stones above the threshold (\%) for OD (Left) and TD (Right) with different data sources provided as input to the model. The x-axis represents the number of stones confidently classified by \gemnet{} with a probability greater than a fixed threshold value (see Fig. \ref{fig:main} (c)). Starting from the left and moving towards the right on the same axis, we gradually increase the threshold and indicate the number of stones that the model confidently classifies for each resulting subset, along with the corresponding accuracy (y-axis). Data sources not present in the legend are masked.}  
    
    \label{fig:sapphire_accuracy_sources}
\end{figure}

\paragraph{Influence of different data sources.}
Fig. \ref{fig:sapphire_accuracy_sources} illustrates the relationship between \gemnet{}'s accuracy and the number of confidently predicted stones for OD (left) and TD (right), where a stone is considered confidently classified by \gemnet{} if the model's probability  associated with its final prediction exceeds the threshold value.
\noindent
The figure reveals a consistent trend across all data sources and tasks: higher levels of confidence in \gemnet{}'s prediction lead to higher accuracy. 
In other words, when the model assigns a high probability to a certain class, it is more likely to be correct. The observed strong correlation between accuracy and confidence in this experiment validates the effectiveness of our confidence-thresholding procedure. \\
The results indicate that, for OD, using ICP-MS data leads to ${\sim} 4\%$ higher accuracy compared to the next best single data source, UV, across the entire range of values on the x-axis. This highlights the high-quality information provided by ICP-MS data for this task. Moreover, combining UV and XRF data sources yields a model with comparable performance to that obtained with ICP-MS data, despite the latter's higher complexity and cost. This suggests that combining standard and less expensive analytical data sources can be as effective as the more expensive ICP-MS method. \\
In TD, while the best results are achieved by integrating UV and FTIR data, \gemnet{} still reaches similar levels of accuracy when only utilizing FTIR data. This is noteworthy since experts typically rely on both data sources to reach a final conclusion in their analysis and suggests that \gemnet{} can serve as a valuable tool even in scenarios where using multiple data sources is not feasible.

\paragraph{Prediction consistency analysis.}

\begin{figure}[H]
    \centering
    \includegraphics[scale = 0.28]{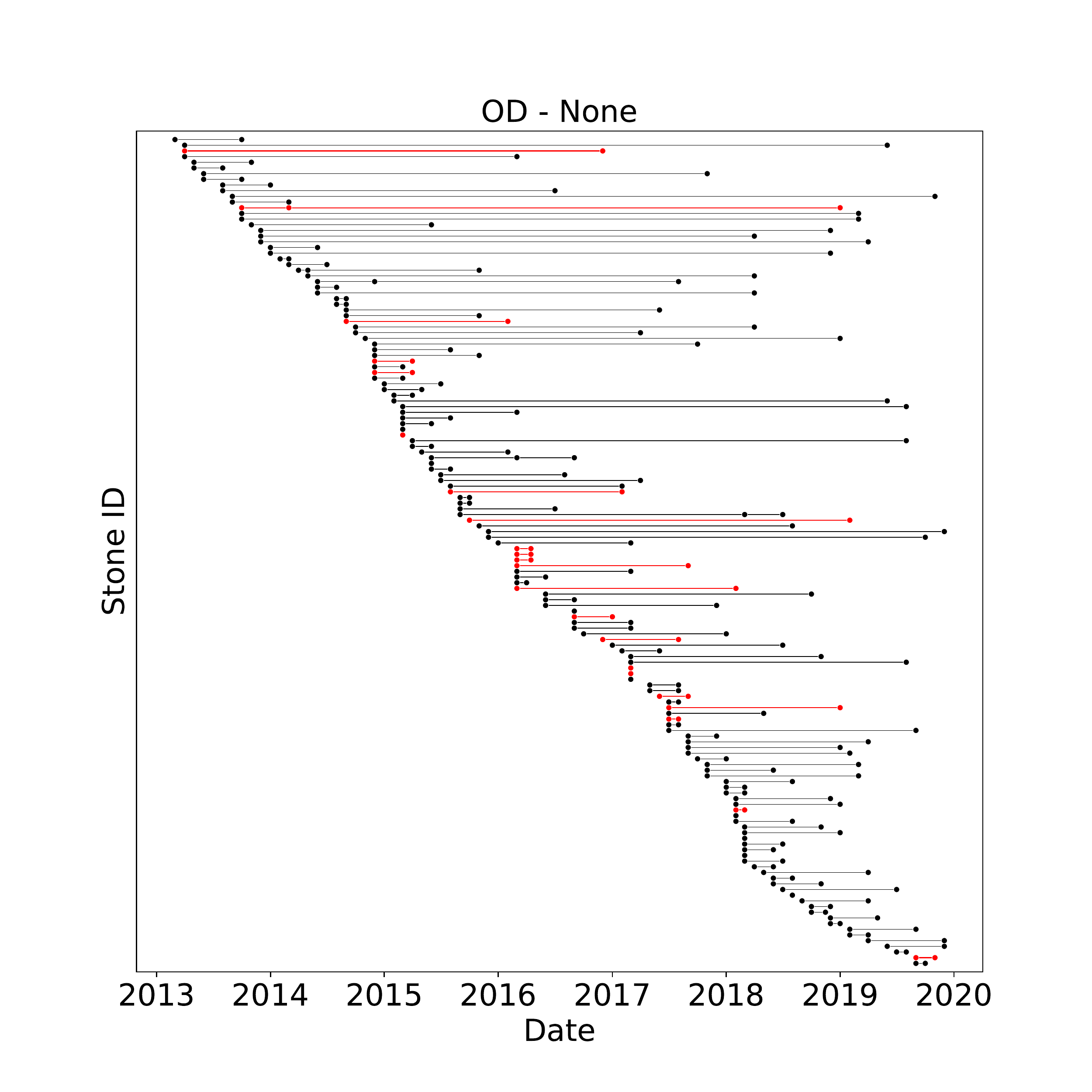} 
    \includegraphics[scale = 0.28]{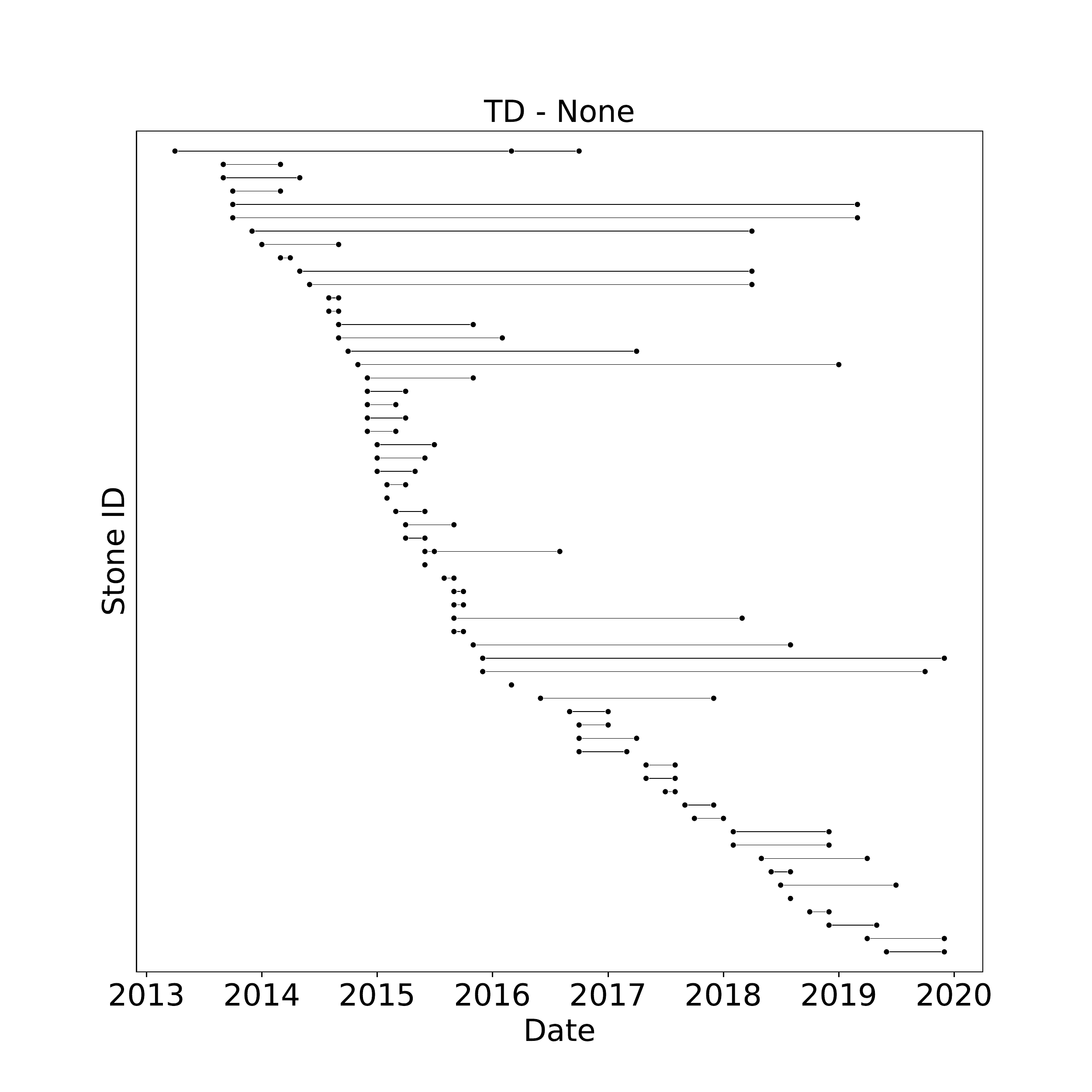} \\
    \includegraphics[scale = 0.28]{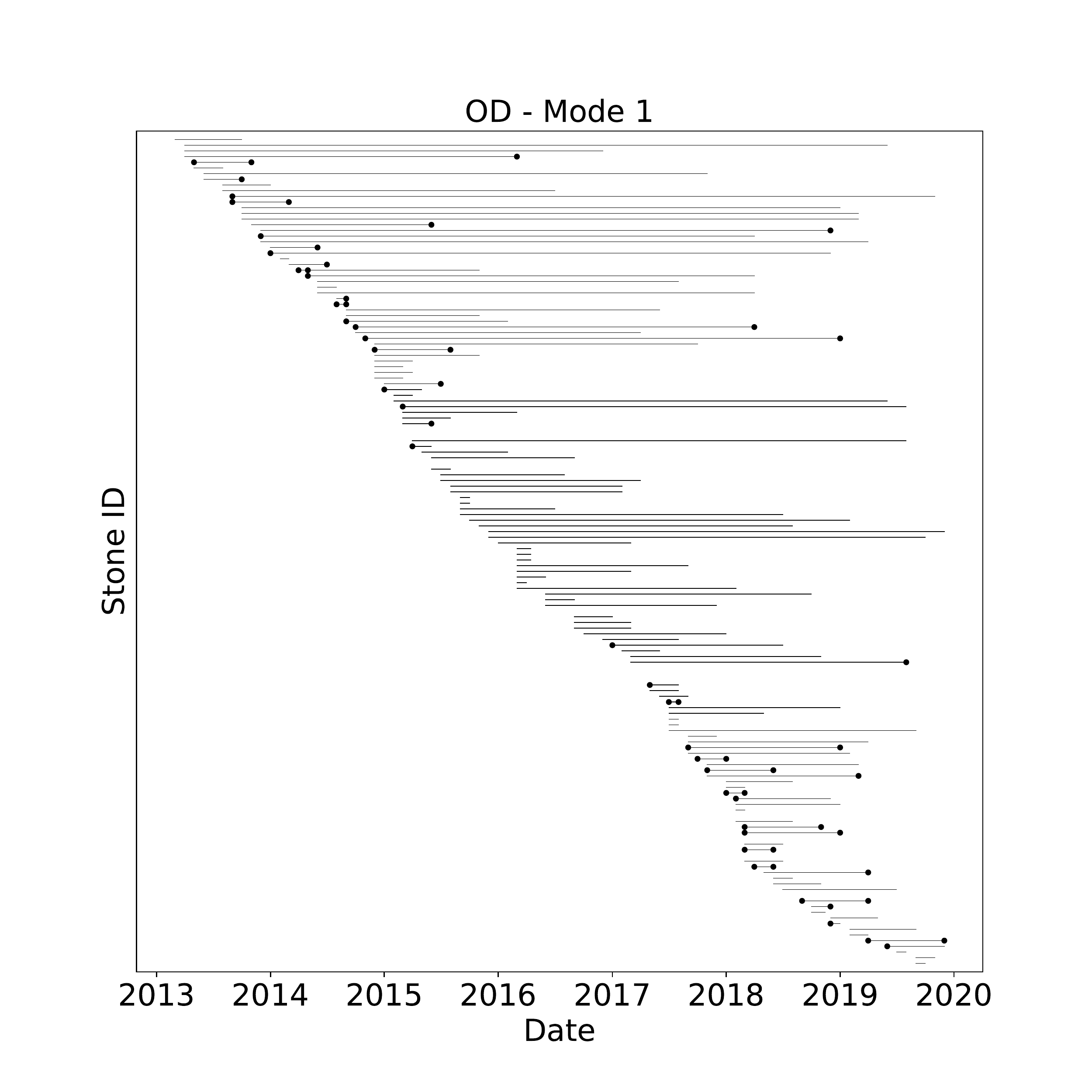} 
    \includegraphics[scale = 0.28]{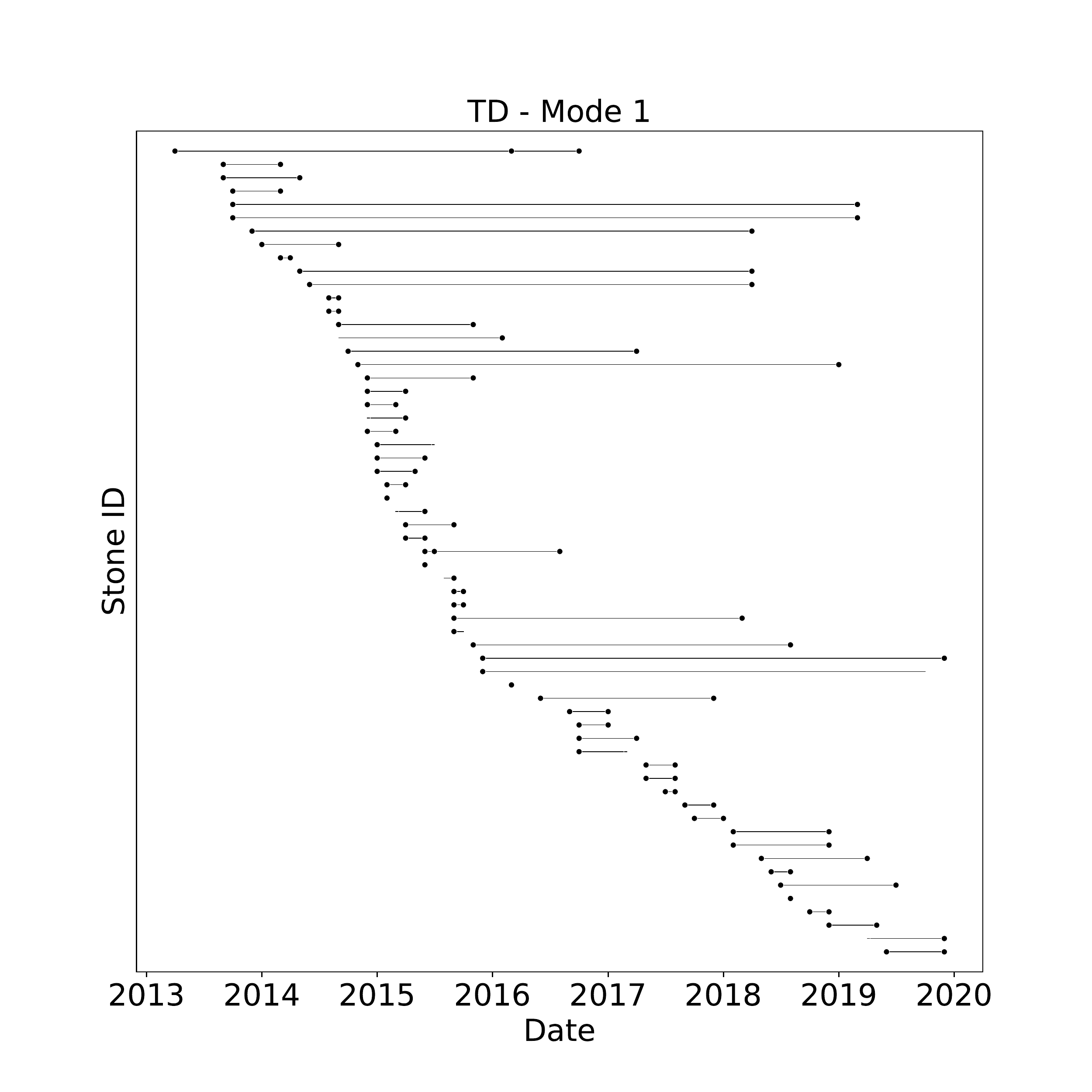}\\ 
    \includegraphics[scale = 0.28]
    {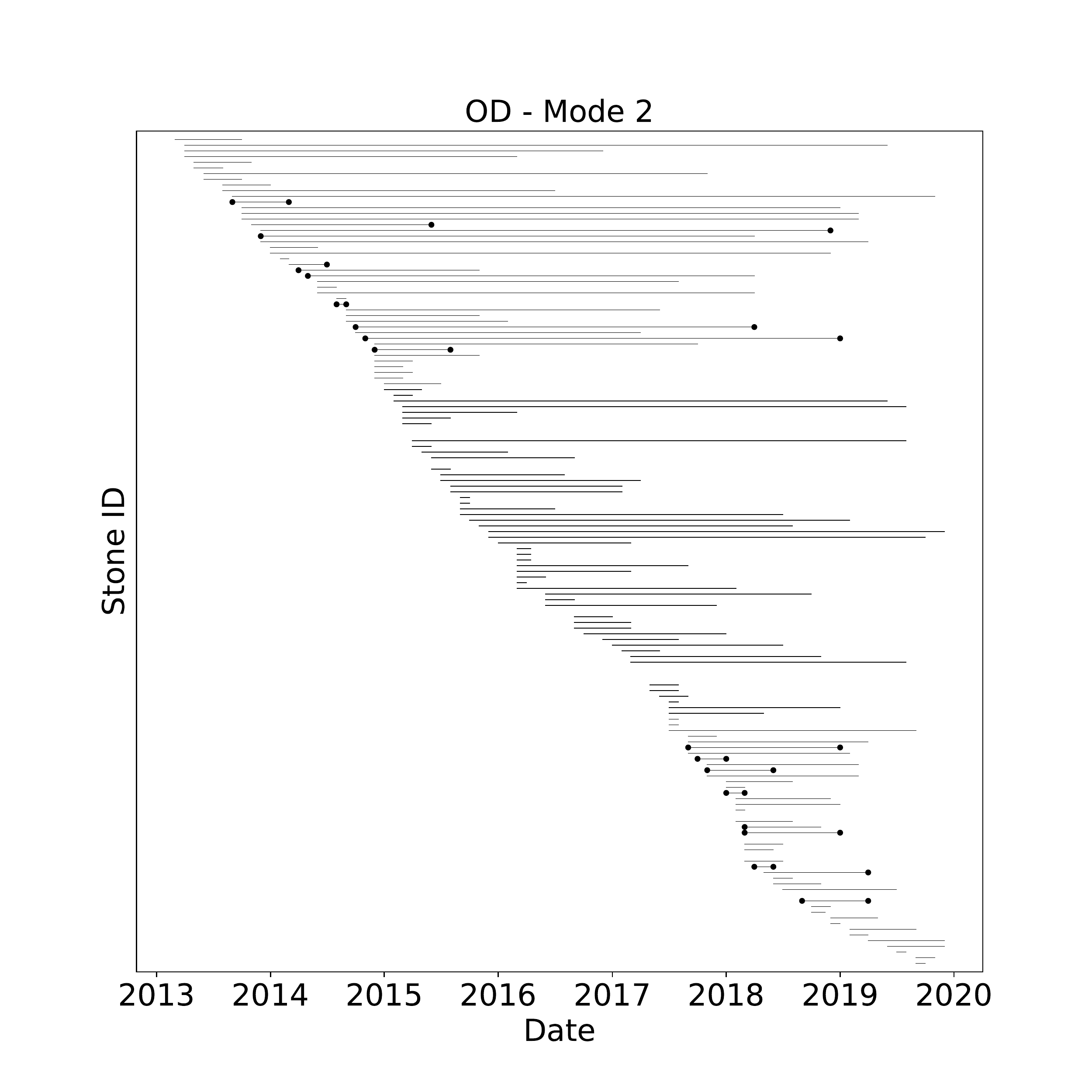} 
    \includegraphics[scale = 0.28]
    {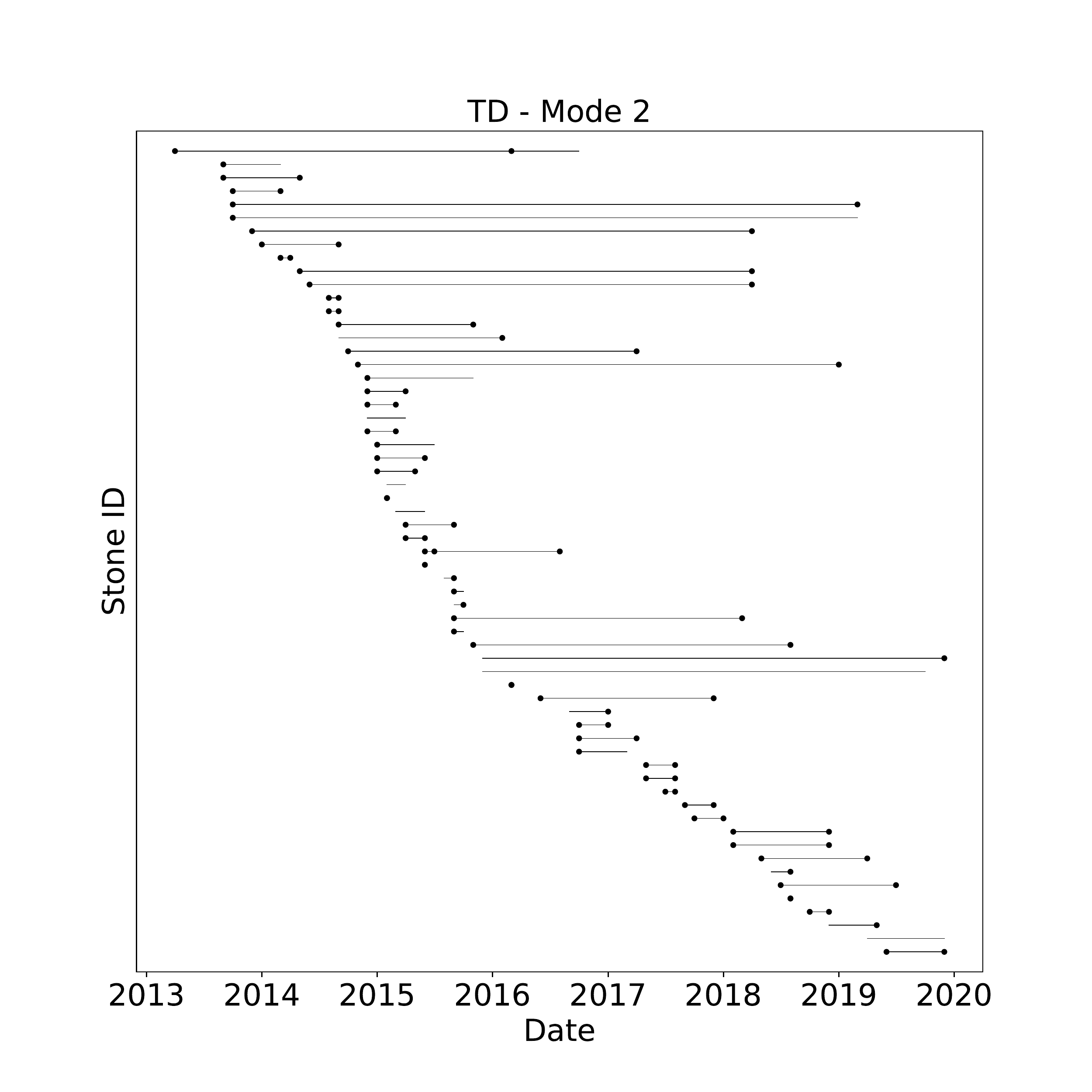} 
    \caption{\gemnet{} predictions over time for OD (Left column) and TD (Right column). Each horizontal line is used to indicate a different stone that is analyzed multiple times over the years. (Upper row) \texttt{None} setup; (Middle row) \texttt{Mode 1} setup; (Bottom row) \texttt{Mode 2} setup. 
    }  
    
    \label{fig:sapphire_consistency}
\end{figure}

To evaluate the accuracy and reliability of \gemnet{}, we analyze if the network generates consistent results for the same gemstone when data is gathered from different instruments, at various times, and under varying conditions. \\
Since gemstones are often subject to multiple analyses during their lifespan, inconsistent evaluations can lead to doubts about the authenticity of the asset, as well as legal and financial complications. Thus, assessing whether the predictions have remained consistent over time is crucial.\\
Fig. \ref{fig:sapphire_consistency} illustrates the predictions of \gemnet{} on the OD (left columns) and TD (right column) tasks for only those gemstones that underwent multiple evaluations over the years, for the \texttt{None} (first row), \texttt{Mode~1} (second row), and \texttt{Mode~2} (third row) model setups. Each stone in our collection that was assessed more than once is represented in each of the six panels by a line connecting the different evaluations. A black line is drawn if the model's predictions are consistent across evaluations, while a red line is drawn if they are not. Dots located at the extremes of the line indicate the predictions of \gemnet{}: the absence of a dot implies that \gemnet{}'s confidence did not exceed the threshold and hence, a decisive conclusion could not be drawn.
It is worth emphasizing that scenarios, where uncertainty prevents a change in prediction (indicated by black lines and no dots), are generally more desirable than inconsistent predictions (shown in red lines). Uncertainty prompts experts to carry out additional analyses to reduce the chances of making mistakes.\\ 
The results displayed in Fig.\ref{fig:sapphire_consistency} demonstrate that even without a threshold (upper row), \gemnet{} exhibits a good level of consistency in its predictions, with only 23 inconsistent predictions out of 148 for OD and zero out of 62 for TD. However, when \gemnet{} is employed in \texttt{Mode~1} and \texttt{Mode~2} setups (middle and bottom rows), all inconsistent predictions disappear as the model's outputs for such cases fall below the threshold, reflecting its uncertainty for those particularly challenging and ambiguous samples.\\
This experiment highlights the value of our confidence-thresholding methodology, which enables the user to disregard predictions that the model is not highly confident about, reducing the risk of incorrect predictions.

\noindent

\section*{Conclusion}
This study introduces \gemnet{}, a novel deep-learning approach for automated origin determination and heat treatment detection of gemstones. \gemnet{} is capable of handling complex and varied data structures and can enhance prediction accuracy by capturing correlations between different data modalities. Its architecture, based on transformers and convolutional neural networks, enables flexible gemstone classification using any combination of diverse data sources and allows for simultaneous end-to-end processing of tabular and spectral data.
\gemnet{} provides numerous benefits. Firstly, its predictions are well calibrated as it outputs correct predictions with high confidence on a large percentage of test samples. This is in contrast to the expert-based evaluation, which provides confident predictions on a significantly smaller subset of stones. Secondly, \gemnet{} provides excellent results by taking as input inexpensive data sources only, hence limiting the reliance on more costly analytic methods, like ICP-MS.\\
Overall, \gemnet{} has the potential to drastically impact the gemstone industry. Its application can result in significant cost savings and can allow human experts to focus on more value-adding activities in the area of research and development. The deployment of \gemnet{} would be crucial in standardizing the gemstone analysis process, significantly reducing the incidence of ambiguities and increasing trust levels in the entire marketplace. In conclusion, we hope that our results, together with the code and data we will make publicly available, will stimulate more investigations in this domain and advance the creation of novel techniques and tools for gemstone analysis automation.

\section*{Methods}

\subsection*{Data Sources}
The following devices and methods were utilized to collect the data used in this study:
\paragraph{ICP-MS.} For ICP-MS data, we used an Elemental Scientific (ESI) 193 nm excimer laser ablation system\footnote{https://www.icpmslasers.com/laserablation/nwr193/} with a large-format sample chamber and a small-volume, flexible cup that collected the ablated material. Three ablations were created for each stone, having a 50-micrometer diameter spot size, 15 Hz repetition rate, and 6 J/cm$^2$ fluence. For each ablation, the materials were conveyed to the ICP via a blend of He (1000 ml/min) and Ar (700 ml/min) gases, where the material got ionized. Finally the ions were transported to an Agilent 8800 mass spectrometer, which measured the following elements / isotopes: $\mathrm{^7Li}$, $\mathrm{^9Be}$, $\mathrm{^{25}Mg}$, $\mathrm{^{27}Al}$, $\mathrm{^{29}Si}$, $\mathrm{^{45}Sc}$, $\mathrm{^{47}Ti}$, $\mathrm{^{49}Ti}$, $\mathrm{^{51}V}$, $\mathrm{^{52}Cr}$, $\mathrm{^{53}Cr}$, $\mathrm{^{55}Mn}$, $\mathrm{^{56}Fe}$, $\mathrm{^{57}Fe}$, $\mathrm{^{59}Co}$, $\mathrm{^{62}Ni}$, $\mathrm{^{71}Ga}$, $\mathrm{^{89}Y}$, $\mathrm{^{90}Zr}$, $\mathrm{^{93}Nb}$, $\mathrm{^{118}Sn}$, $\mathrm{^{140}Ce}$, $\mathrm{^{146}Nd}$, $\mathrm{^{176}Hf}$, $\mathrm{^{181}Ta}$, $\mathrm{^{193}Ir}$, $\mathrm{^{195}Pt}$. The acquired data was then processed from counts/second to concentration using Glitter \cite{griffin2008glitter}, with NIST 612 \cite{jochum2011determination} as the primary calibration standard and BHVO-2G \cite{jochum2005chemical} and ATHO-G \cite{jochum2006mpi} as secondary standards. A value of 99 wt\% $\mathrm{Al_2O_3}$ was used as an internal standard for all corundum. Following gemological-driven analysis, we focussed for our study on the following entries: $\mathrm{^9Be}$, $\mathrm{^{25}Mg}$, $\mathrm{^{27}Al}$, $\mathrm{^{45}Sc}$, $\mathrm{^{49}Ti}$, $\mathrm{^{51}V}$, $\mathrm{^{53}Cr}$, $\mathrm{^{57}Fe}$, $\mathrm{^{62}Ni}$, $\mathrm{^{71}Ga}$, $\mathrm{^{90}Zr}$, $\mathrm{^{118}Sn}$, $\mathrm{^{140}Ce}$, $\mathrm{^{146}Nd}$, $\mathrm{^{176}Hf}$, $\mathrm{^{181}Ta}$. 

\paragraph{FTIR.} Non-polarized FTIR spectra were collected in air using a Varian 640 FTIR spectrometer equipped with a KBr beam splitter and a deuterated triglycene sulfate (DTGS) detector. For each sample, three measurements in perpendicular directions were conducted either using diffuse reflectance (DRIFT) or with transmitted light. For each measurement, a total of 64 scans with a resolution of 1 cm$^{-1}$ to 4 cm$^{-1}$ were collected and averaged. This was done for the wavenumber range of 200 cm$^{-1}$ to 7000 cm$^{-1}$, with a background collected at regular intervals.
As the measurements had different intervals and offsets due to differences in software version and settings, we homogenized the data so that every spectrum had a step size of 1 cm$^{-1}$. This was done by a cubic spline interpolation on the available data. As not all data were collected over the range of 200 cm$^{-1}$ to 7000 cm$^{-1}$, we padded the missing values with zeroes. Further, any spectra which had a measurement with a value smaller than -5 or greater than 10 were dropped as these values were extreme outliers and not in the expected range of the measurement. This filtering reduced the data set by less than 1\%. The resulting spectral data consisted of 6801 data points per measurement.

\paragraph{XRF.} XRF (ED-XRF) measurements of major, minor, and trace elements were conducted using a Thermo Fisher Scientific QUANTX, with a silicon drift detector (SDD), 1 mm collimator, and over an applied energy range of 4-50 kV, with a variety of filters used to reduce spectral interferences on critical elements. For blue sapphires, the only minor and trace elements consistently detectable are Ti, Cr, V, Fe, and Ga, along with the major element Al. In order to identify treatments or synthetic samples, Pb, W, and Pt were included during most blue sapphire measurement routines. For blue sapphires, we discarded the ED-XRF measurement, if
    \begin{itemize}
        \item The $\mathrm{Fe_2O_3}$ value is above 40'000~ppm
        \item The $\mathrm{Al_2O_3}$ value is under 850'000~ppm
        \item The $\mathrm{Cr_2O_3}$ value is above 10'000~ppm
        \item The $\mathrm{TiO_2}$ value is above 6'000~ppm
    \end{itemize}
Such outliers do occur occasionally in XRF measurements due to various reasons such as diffraction peaks induced by the crystal structure of the minerals. ED-XRF data is tabular in nature, having 26 entries describing the concentration of certain chemical compounds.

\paragraph{UV.} Polarised UV (UV-Vis-NIR) spectra were collected using a Varian (now Agilent) Cary~5000, using deuterium and tungsten halogen light sources and indium gallium arsenide (InGaAs) detector. Measurements were performed over the wavelength range 280-880~nm with a step size of 0.5~nm, using both a reference and sample line equipped with polarisers and beam condensers. In most cases, two measurements in perpendicular polarisations were taken on each sample. In the case of single measurements, the measurement was duplicated to be consistent with the bi-polar measurements. As the absorbance can not be negative, any spectra with negative values were discarded. This could occur due to faulty measurements. The resulting final data sample consisted of 2~x~1201 entries.\\
    
\noindent
During the time period from which the data included in this study were obtained, several variants of these instruments were used and a minority of data were collected on other instruments, not detailed here. Data consistency between these different models was maintained through standardized acquisition protocols and the use of identical calibration and secondary reference materials.

\subsection*{\gemnet{} Architecture}
\gemnet{} is an artificial neural network created to process multi-modal data from gemological laboratories. It is composed of a UV encoder, an FTIR encoder, and a single elemental analysis encoder that processes XRF (and optionally ICP-MS jointly) data. The encoders generate embeddings which are then combined by the network's head. This head comprises a concatenation layer to combine the encoders' outputs, batch normalization, and a final linear classification layer.\\ 
The UV and FTIR encoders are strided convolutional neural networks with skip connections as proposed in \cite{ho2019rapid}. At the core of their architecture, there are six residual connection layers, each with a hidden dimension of 128, kernel size of 17, and strides of 2. These blocks are preceded by a first convolution layer of kernel size 59. For UV measurements, which involve two spectra taken in perpendicular directions, the input channel of the first convolution layer has a dimension of two, while for FTIR measurements, a single spectrum is used and the first convolution has a dimension of one.
After the skip connection blocks, the FTIR and UV encoders employ a single convolution channel mapping the hidden dimension from 128 to 1, resulting in final embeddings of length 213 and 190 respectively.
The parameter selection was based on a  preliminary grid search. Particularly, we found that a smaller kernel size or fewer residual connection blocks caused a decrease in performance, while larger dimensions resulted in high memory usage and slow training with no increase in accuracy. 
The elemental analysis encoder is based on the \texttt{SAINT} framework introduced by \cite{somepalli2021saint}, which was specifically designed to provide a sample-efficient deep learning method for tabular data. We opted to follow the \texttt{Both} configuration of the original paper, which deploys both intrasample and intersample attention mechanisms. Intrasample attention is a standard self-attention mechanism, operating on input features (rows), while intersample attention compares specific input features across different samples (columns). Our implementation follows the same hyper-parameters as described in the original paper. However, since our setting only has one sample at inference time rather than a batch, we decided to pre-append a series of reference stones to the batch for both training and testing to avoid a shift in the distribution caused by the intrasample mechanism.
XRF and ICP-MS data are concatenated before the encoder allowing for the model to learn dependencies between both data types. The output tensor of the elemental analysis encoder has a single hidden dimension with a length of~32. \\
Finally, the head of the network is composed of a concatenation layer, followed by a batch normalization and a readout layer. The concatenation layer combines the one-dimensional embeddings from the UV, FTIR, and elemental analysis encoders into a single tensor by concatenating them along the time dimension. This tensor is then fed into the batch normalization and classification layers. The readout layer is composed of a linear layer with a softmax activation function, and its output is the final classification probability for each class.

\subsection*{Training and Testing \gemnet{}}
\subsubsection*{Training details}
For training our model, we randomly partitioned the training data into 80\% for training, and 20\% for validation, saving the model's weights every 5 epochs during the 250 epochs. We then picked the best model in terms of accuracy from the saved weights. During training the batch size was set to 16 and the learning rate was set to 0.0001, with an automatic decay of factor 10 if there was no improvement for more than 10 epochs. To allow the model to learn to handle missing data, we randomly masked one data source with a probability of 0.7 during training, replacing the values with the mean value across the dataset. We did not perform any data normalization or augmentation, as it was found to be detrimental in early experiments. We repeated the same procedure for each of the five folds in the cross-validation procedure. In each fold, the test data was not used, neither for training nor validation.\\
In order to generate the final results from the folds, we followed the procedures laid out in \cite{forman2010apples}. Specifically, for all results apart from Fig. \ref{fig:sapphire_accuracy_sources}, we concatenated the predictions of \gemnet{} from each fold and then calculated the final statistic. For Fig. \ref{fig:sapphire_accuracy_sources}  we first computed the curve in each fold, and then obtained the final curve by computing the average of the curves from each fold.
We conducted all the experiments on a machine equipped with an NVIDIA GeForce RTX 2080 Ti with 12 CPU cores.

\subsubsection*{Confidence-thresholding procedure}\label{Cal procedure}
The purpose of our confidence-thresholding procedure is to determine the reliability of the model's prediction based on the associated level of confidence. More formally, we define the model's confidence $c$ for a given prediction $p$ as the maximum value of the last softmax layer. A reliable prediction is defined as a prediction for which $c$ is greater than some predefined threshold $\hat{c}$ (e.g., 0.95). To determine the value of $\hat{c}$, we perform the following steps: first, we compute the model’s predictions $\{p_{i}\}_{i=1}^N$ and associated confidence values $\{c_{i}\}_{i=1}^N$ for each stone in the training set after training. Then, we sort the stones by confidence values from lowest to highest (i.e. $c_{(1)} \leq c_{(2)} \leq ... \leq c_{(N)}$). Next, we iteratively compute the accuracy of the subset of stones with the least confidence removed until the subset accuracy is greater than a pre-specified value $\epsilon$ (e.g. 95\%). We define the accuracy of the subset of stones corresponding to the entire dataset minus the $k$ stones with the smallest confidence values as $\mathcal{A}_{N-k}$. At inference time, the threshold of the least confident stone in this subset is used to decide if a model prediction is reliable. Specifically, we set $\hat{c}=c_{k^*}$ where
where $k^*$ is defined as the index that satisfies $\mathcal{A}_{N-k'} \geq \epsilon$ for the first time, where $k' \in {1,2,\dots,N}$.

\FloatBarrier

\bibliographystyle{ieeetr}
\bibliography{references}

\begin{thebibliography}{10}

\bibitem{phichaikamjornwut2019conclusive}
B.~Phichaikamjornwut, S.~Pongkrapan, S.~Intarasiri, and D.~Bootkul,
  ``Conclusive comparison of gamma irradiation and heat treatment for color
  enhancement of rubellite from mozambique,'' {\em Vibrational Spectroscopy},
  vol.~103, p.~102926, 2019.

\bibitem{jaliya2020characterization}
R.~Jaliya, P.~Dharmaratne, and K.~Wijesekara, ``Characterization of heat
  treated geuda gemstones for different furnace conditions using ftir, xrd and
  uv--visible spectroscopy methods,'' {\em Solid Earth Sciences}, vol.~5,
  no.~4, pp.~282--289, 2020.

\bibitem{emmett2003beryllium}
J.~L. Emmett, K.~Scarratt, S.~F. McClure, T.~Moses, T.~R. Douthit, R.~Hughes,
  S.~Novak, J.~E. Shigley, W.~Wang, O.~Bordelon, {\em et~al.}, ``Beryllium
  diffusion of ruby and sapphire,'' {\em Gems \& Gemology}, vol.~39, no.~2,
  pp.~84--135, 2003.

\bibitem{johnson1999identification}
M.~L. Johnson, S.~Elen, and S.~Muhlmeister, ``On the identification of various
  emerald filling substances,'' {\em Gems \& Gemology}, vol.~35, no.~2,
  pp.~82--107, 1999.

\bibitem{pardieu2020field}
V.~Pardieu, ``Field gemology,'' {\em The evolution of data collection.
  InColor}, vol.~46, pp.~100--106, 2020.

\bibitem{dissanayake1999sri}
C.~Dissanayake and R.~Chandrajith, ``Sri lanka--madagascar gondwana linkage:
  evidence for a pan-african mineral belt,'' {\em The Journal of Geology},
  vol.~107, no.~2, pp.~223--235, 1999.

\bibitem{rankin2006gubelin}
A.~Rankin, ``Gubelin, ej and koivula, ji photoatlas of inclusions in gemstones,
  volume 2. basel (opinio publishers), hardback, 2005, 829 pp. isbn
  3-03999029-2, price us \$260.,'' 2006.

\bibitem{hughes2019madagascar}
E.~B. Hughes and R.~Perkins, ``Madagascar sapphire: Low-temperature heat
  treatment experiments,'' {\em Gems Gemol}, vol.~55, pp.~184--196, 2019.

\bibitem{groat2019review}
L.~A. Groat, G.~Giuliani, J.~Stone-Sundberg, Z.~Sun, N.~D. Renfro, and A.~C.
  Palke, ``A review of analytical methods used in geographic origin
  determination of gemstones.,'' {\em Gems \& Gemology}, vol.~55, no.~4, 2019.

\bibitem{scarani2017gemological}
A.~Scarani and M.~{\AA}str{\"o}m, ``Gemological applications of uv-vis-nir
  spectroscopy,'' {\em Riv. Ital. Gemmol}, vol.~7, pp.~1--9, 2017.

\bibitem{kitawaki2006identification}
H.~Kitawaki, A.~Abduriyim, and M.~Okano, ``Identification of heat-treated
  corundum.,'' {\em Gems \& Gemology}, vol.~42, no.~3, 2006.

\bibitem{joseph2000characterization}
D.~Joseph, M.~Lal, P.~Shinde, and B.~Padalia, ``Characterization of gem stones
  (rubies and sapphires) by energy-dispersive x-ray fluorescence
  spectrometry,'' {\em X-Ray Spectrometry: An International Journal}, vol.~29,
  no.~2, pp.~147--150, 2000.

\bibitem{guillong2001quasi}
M.~Guillong and D.~G{\"u}nther, ``Quasi ‘non-destructive’laser
  ablation-inductively coupled plasma-mass spectrometry fingerprinting of
  sapphires,'' {\em Spectrochimica Acta Part B: Atomic Spectroscopy}, vol.~56,
  no.~7, pp.~1219--1231, 2001.

\bibitem{rossman2009geochemistry}
G.~R. Rossman, ``The geochemistry of gems and its relevance to gemology:
  Different traces, different prices,'' {\em Elements}, vol.~5, no.~3,
  pp.~159--162, 2009.

\bibitem{palke2019geographic}
A.~C. Palke, S.~Saeseaw, N.~D. Renfro, Z.~Sun, and S.~F. McClure, ``Geographic
  origin determination of blue sapphire.,'' {\em Gems \& Gemology}, vol.~55,
  no.~4, 2019.

\bibitem{matsci1}
R.~Ramprasad, R.~Batra, G.~Pilania, A.~Mannodi-Kanakkithodi, and C.~Kim,
  ``Machine learning in materials informatics: recent applications and
  prospects,'' {\em npj Computational Materials}, vol.~3, no.~1, p.~54, 2017.

\bibitem{matsci2}
L.~Ward, A.~Agrawal, A.~Choudhary, and C.~Wolverton, ``A general-purpose
  machine learning framework for predicting properties of inorganic
  materials,'' {\em npj Computational Materials}, vol.~2, no.~1, pp.~1--7,
  2016.

\bibitem{matsci3}
P.~Raccuglia, K.~C. Elbert, P.~D. Adler, C.~Falk, M.~B. Wenny, A.~Mollo,
  M.~Zeller, S.~A. Friedler, J.~Schrier, and A.~J. Norquist,
  ``Machine-learning-assisted materials discovery using failed experiments,''
  {\em Nature}, vol.~533, no.~7601, pp.~73--76, 2016.

\bibitem{clisci1}
S.~Rasp, M.~S. Pritchard, and P.~Gentine, ``Deep learning to represent subgrid
  processes in climate models,'' {\em Proceedings of the National Academy of
  Sciences}, vol.~115, no.~39, pp.~9684--9689, 2018.

\bibitem{clisci2}
L.~Zhang, L.~Zhang, and B.~Du, ``Deep learning for remote sensing data: A
  technical tutorial on the state of the art,'' {\em IEEE Geoscience and Remote
  Sensing Magazine}, vol.~4, pp.~22--40, 2016.

\bibitem{dg1}
J.~Vamathevan, D.~Clark, P.~Czodrowski, I.~Dunham, E.~Ferran, G.~Lee, B.~Li,
  A.~Madabhushi, P.~Shah, M.~Spitzer, {\em et~al.}, ``Applications of machine
  learning in drug discovery and development,'' {\em Nature reviews Drug
  discovery}, vol.~18, no.~6, pp.~463--477, 2019.

\bibitem{dg2}
G.~B. Goh, N.~O. Hodas, and A.~Vishnu, ``Deep learning for computational
  chemistry,'' {\em Journal of computational chemistry}, vol.~38, no.~16,
  pp.~1291--1307, 2017.

\bibitem{dg3}
H.~Chen, O.~Engkvist, Y.~Wang, M.~Olivecrona, and T.~Blaschke, ``The rise of
  deep learning in drug discovery,'' {\em Drug discovery today}, vol.~23,
  no.~6, pp.~1241--1250, 2018.

\bibitem{chow2021automatic}
B.~H.~Y. Chow and C.~C. Reyes-Aldasoro, ``Automatic gemstone classification
  using computer vision,'' {\em Minerals}, vol.~12, no.~1, p.~60, 2021.

\bibitem{botejuresearch}
W.~Boteju, M.~Gunarathna, T.~Fonseka, H.~Peiris, L.~Silva, N.~Jayathilaka,
  L.~Jayasekara, P.~Perera, H.~Jayaweera, M.~Gunewardene, {\em et~al.},
  ``Research article a quantitative approach to gemstone identification using
  raman spectroscopy combined with machine learning,''

\bibitem{ostreika2021classification}
A.~Ostreika, M.~Pivoras, A.~Misevi{\v{c}}ius, T.~Skersys, and L.~Paulauskas,
  ``Classification of objects by shape applied to amber gemstone
  classification,'' {\em Applied Sciences}, vol.~11, no.~3, p.~1024, 2021.

\bibitem{qiu2019feasibility}
J.-T. Qiu, L.~Qiu, and H.-X. Mu, ``Feasibility of gem identification using
  reflectance spectra coupled with artificial intelligence,'' {\em Spectroscopy
  Letters}, vol.~52, no.~9, pp.~520--532, 2019.

\bibitem{wang2015automated}
D.~Wang, L.~Bischof, R.~Lagerstrom, V.~Hilsenstein, A.~Hornabrook, and
  G.~Hornabrook, ``Automated opal grading by imaging and statistical
  learning,'' {\em IEEE transactions on systems, man, and cybernetics:
  systems}, vol.~46, no.~2, pp.~185--201, 2015.

\bibitem{amarasekara2021convolutional}
S.~Amarasekara and R.~Meegama, ``Convolutional neural network for
  classification and value estimation of selected gemstones in sri lanka,''
  2021.

\bibitem{shukla2022artificial}
A.~K. Shukla, {\em Artificial Intelligence and Spectroscopic Techniques for
  Gemology Applications}.
\newblock IOP Publishing, 2022.

\bibitem{dumoulin2016guide}
V.~Dumoulin and F.~Visin, ``A guide to convolution arithmetic for deep
  learning,'' {\em arXiv preprint arXiv:1603.07285}, 2016.

\bibitem{vaswani2017attention}
A.~Vaswani, N.~Shazeer, N.~Parmar, J.~Uszkoreit, L.~Jones, A.~N. Gomez,
  {\L}.~Kaiser, and I.~Polosukhin, ``Attention is all you need,'' {\em Advances
  in neural information processing systems}, vol.~30, 2017.

\bibitem{ho2019rapid}
C.-S. Ho, N.~Jean, C.~A. Hogan, L.~Blackmon, S.~S. Jeffrey, M.~Holodniy,
  N.~Banaei, A.~A. Saleh, S.~Ermon, and J.~Dionne, ``Rapid identification of
  pathogenic bacteria using raman spectroscopy and deep learning,'' {\em Nature
  communications}, vol.~10, no.~1, pp.~1--8, 2019.

\bibitem{somepalli2021saint}
G.~Somepalli, M.~Goldblum, A.~Schwarzschild, C.~B. Bruss, and T.~Goldstein,
  ``Saint: Improved neural networks for tabular data via row attention and
  contrastive pre-training,'' {\em arXiv preprint arXiv:2106.01342}, 2021.

\bibitem{pyzer2022accelerating}
E.~O. Pyzer-Knapp, J.~W. Pitera, P.~W. Staar, S.~Takeda, T.~Laino, D.~P.
  Sanders, J.~Sexton, J.~R. Smith, and A.~Curioni, ``Accelerating materials
  discovery using artificial intelligence, high performance computing and
  robotics,'' {\em npj Computational Materials}, vol.~8, no.~1, pp.~1--9, 2022.

\bibitem{vaucher2020automated}
A.~C. Vaucher, F.~Zipoli, J.~Geluykens, V.~H. Nair, P.~Schwaller, and T.~Laino,
  ``Automated extraction of chemical synthesis actions from experimental
  procedures,'' {\em Nature communications}, vol.~11, no.~1, p.~3601, 2020.

\bibitem{vaucher2021inferring}
A.~C. Vaucher, P.~Schwaller, J.~Geluykens, V.~H. Nair, A.~Iuliano, and
  T.~Laino, ``Inferring experimental procedures from text-based representations
  of chemical reactions,'' {\em Nature communications}, vol.~12, no.~1,
  pp.~1--11, 2021.

\bibitem{naugler2019automation}
C.~Naugler and D.~L. Church, ``Automation and artificial intelligence in the
  clinical laboratory,'' {\em Critical reviews in clinical laboratory
  sciences}, vol.~56, no.~2, pp.~98--110, 2019.

\bibitem{giuliani2019geology}
G.~Giuliani, L.~A. Groat, {\em et~al.}, ``Geology of corundum and emerald gem
  deposits,'' {\em Gems \& Gemology}, vol.~55, 2019.

\bibitem{smith1995contribution}
C.~P. Smith {\em et~al.}, ``A contribution to understanding the infrared
  spectra of rubies from mong hsu, myanmar,'' {\em Journal of Gemmology},
  vol.~24, no.~5, pp.~321--335, 1995.

\bibitem{gubelin1953inclusions}
E.~J. G{\"u}belin, {\em Inclusions as a Means of Gemstone Identification: With
  258 Illus}.
\newblock Gemological Institute of America, 1953.

\bibitem{griffin2008glitter}
W.~Griffin, ``Glitter: data reduction software for laser ablation icp-ms,''
  {\em Laser Ablation ICP-MS in the Earth Sciences: Current practices and
  outstanding issues}, pp.~308--311, 2008.

\bibitem{jochum2011determination}
K.~P. Jochum, U.~Weis, B.~Stoll, D.~Kuzmin, Q.~Yang, I.~Raczek, D.~E. Jacob,
  A.~Stracke, K.~Birbaum, D.~A. Frick, {\em et~al.}, ``Determination of
  reference values for nist srm 610--617 glasses following iso guidelines,''
  {\em Geostandards and Geoanalytical Research}, vol.~35, no.~4, pp.~397--429,
  2011.

\bibitem{jochum2005chemical}
K.~P. Jochum, M.~Willbold, I.~Raczek, B.~Stoll, and K.~Herwig, ``Chemical
  characterisation of the usgs reference glasses gsa-1g, gsc-1g, gsd-1g,
  gse-1g, bcr-2g, bhvo-2g and bir-1g using epma, id-tims, id-icp-ms and
  la-icp-ms,'' {\em Geostandards and Geoanalytical Research}, vol.~29, no.~3,
  pp.~285--302, 2005.

\bibitem{jochum2006mpi}
K.~P. Jochum, B.~Stoll, K.~Herwig, M.~Willbold, A.~W. Hofmann, M.~Amini,
  S.~Aarburg, W.~Abouchami, E.~Hellebrand, B.~Mocek, {\em et~al.}, ``Mpi-ding
  reference glasses for in situ microanalysis: New reference values for element
  concentrations and isotope ratios,'' {\em Geochemistry, Geophysics,
  Geosystems}, vol.~7, no.~2, 2006.

\bibitem{forman2010apples}
G.~Forman and M.~Scholz, ``Apples-to-apples in cross-validation studies:
  pitfalls in classifier performance measurement,'' {\em Acm Sigkdd
  Explorations Newsletter}, vol.~12, no.~1, pp.~49--57, 2010.

\bibitem{krebs2020evaluation}
M.~Y. Krebs, M.~F. Hardman, D.~G. Pearson, Y.~Luo, A.~J. Fagan, and C.~Sarkar,
  ``An evaluation of the potential for determination of the geographic origin
  of ruby and sapphire using an expanded trace element suite plus sr--pb
  isotope compositions,'' {\em Minerals}, vol.~10, no.~5, p.~447, 2020.

\bibitem{jaegle2021perceiver}
A.~Jaegle, F.~Gimeno, A.~Brock, A.~Zisserman, O.~Vinyals, and J.~Carreira,
  ``Perceiver: General perception with iterative attention,'' 2021.

\end{thebibliography}

\FloatBarrier
\newpage

\clearpage
\pagenumbering{arabic}
\renewcommand*{\thepage}{\arabic{page}}
\appendix

\begin{center}
\huge{\textbf{\gemnet: Accelerating Gemstone classification with Deep Learning: Supplementary Material}}\\
%\vspace{1cm}
%\normalsize
%Tommaso Bendinelli$^{1}$, Luca Biggio$^{1,2}$, Daniel Nyfeler$^{3}$, Abhigyan Ghosh$^{3}$, Peter Tollan$^{3}$, Moritz Alexander Kirschmann$^{1}$ and Olga Fink$^{4}$\\
%\vspace{0.5cm}
%$^1${CSEM SA, Alpnach, Switzerland}\\
%$^2${Data Analytics Lab, ETH, Zürich, Switzerland}\\
%$^3${Gübelin Gem Lab Ltd, Lucerne, Switzerland}\\
%$^4${Laboratory for Intelligent Maintenance and Operations Systems, EPFL, Switzerland}\\
\end{center}

\setcounter{figure}{0}  
\captionsetup[figure]{labelformat={default},labelsep=period,name={Supplementary Fig.}}

\setcounter{table}{0}  
\captionsetup[table]{labelformat={default},labelsep=period,name={Supplementary Table}}

\section*{Supplementary Notes}

\subsection*{Supplementary Note 1: Preliminary experiments}
\begin{figure}[H]
    \centering
    
    \includegraphics[scale = 0.45]{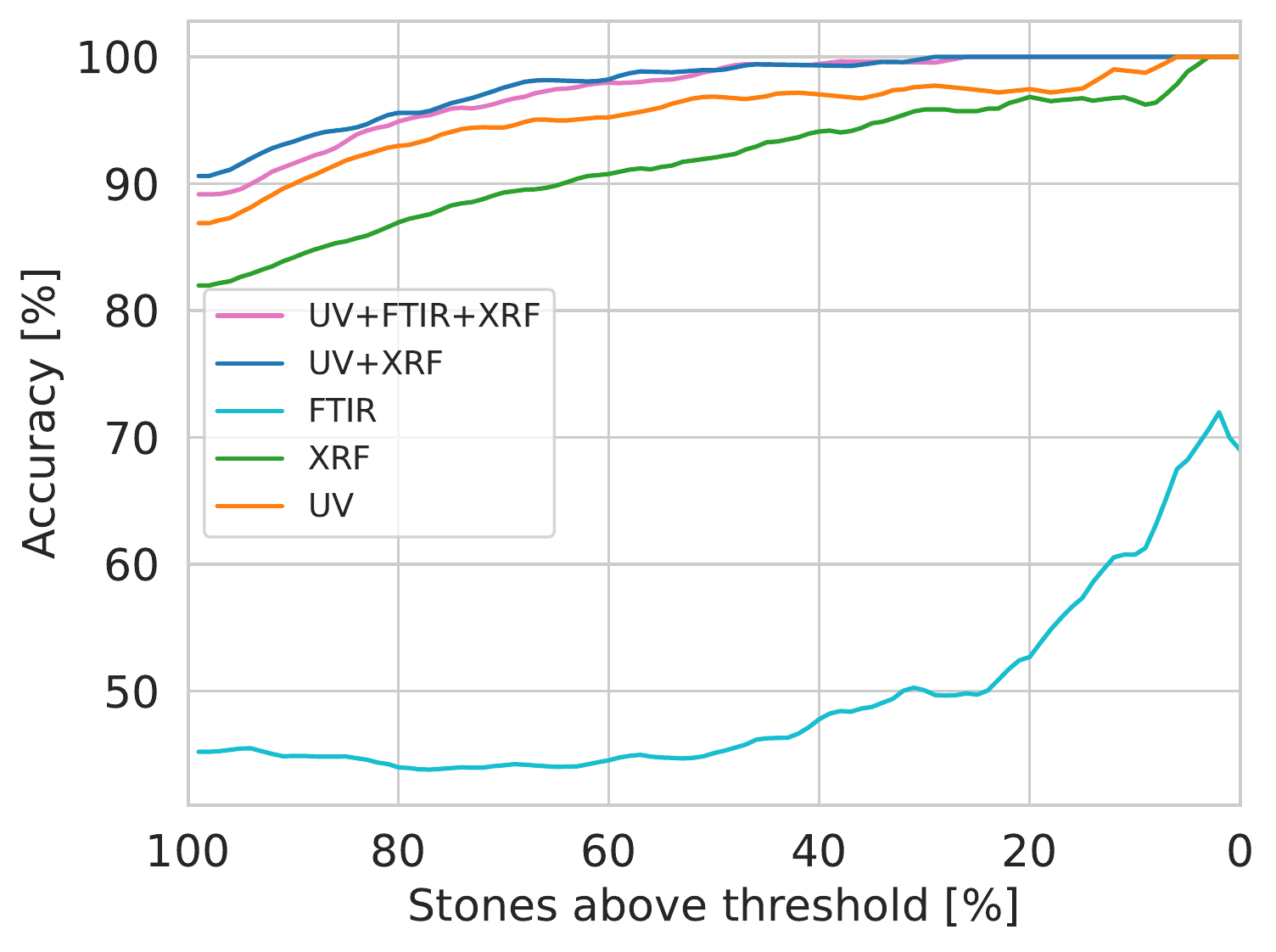} 
    
    \caption{Accuracy (\%) vs. stones above the threshold (\%) for OD with different data sources provided as input to the model. FTIR performs sub-optimally compared to other data sources. Data sources that are not present in the legend are masked.    
    }  
    
    \label{fig:sapphire_accuracy_ablation_FTIR}
\end{figure}

\begin{figure}[H]
    \centering

    \includegraphics[scale = 0.45]{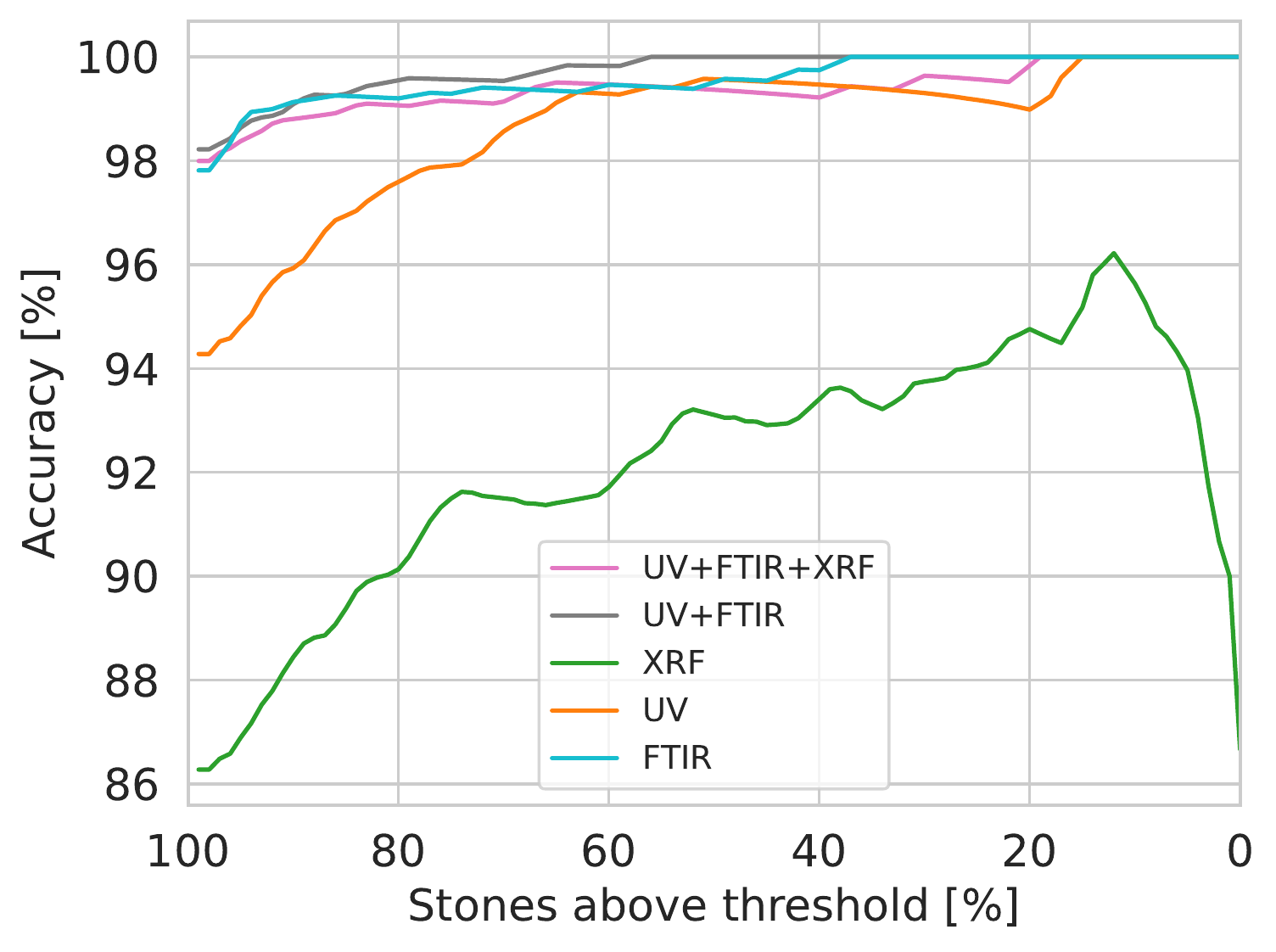} 
    \includegraphics[scale = 0.45]{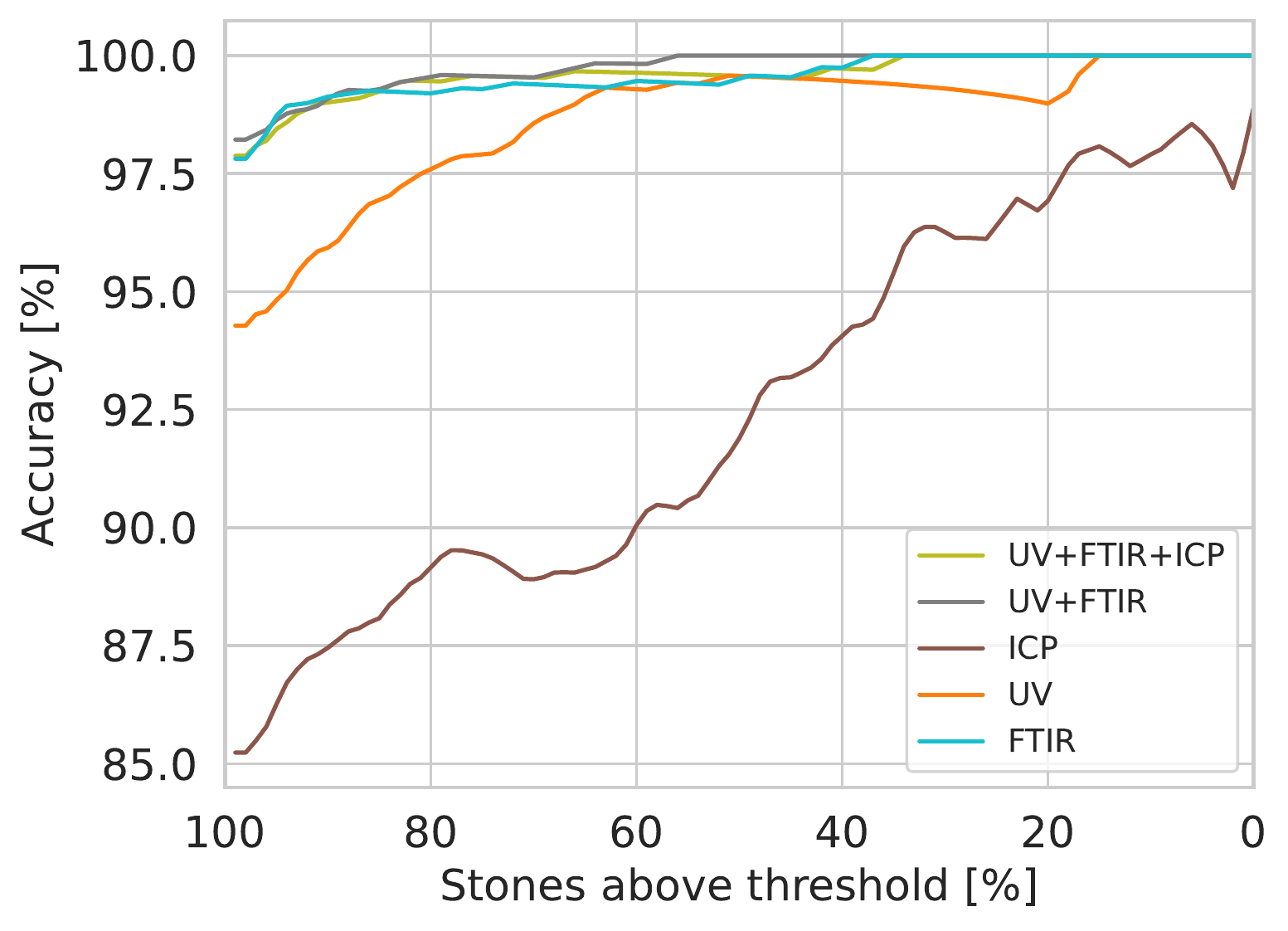} 
    
    \caption{Accuracy [\%] vs. stones above the threshold [\%] for TD with XRF (Left)  and ICP (right). Both XRF and ICP perform sub-optimally compared to other data sources. Data sources that are not present in the legend are masked.   
    }
    
    \label{fig:sapphire_accuracy_ablation_ICP}
\end{figure}

In this Note, we present the performance of \gemnet{} for OD using FTIR, and for TD using XRF and ICP-MS. Supplementary Fig.~\ref{fig:sapphire_accuracy_ablation_FTIR} demonstrates that FTIR does not provide any additional predictive power when combined with other data sources, which is consistent with laboratory practice, as FTIR does not contribute significantly to determination of origin of sapphires. Regarding TD, Supplementary Fig.\ref{fig:sapphire_accuracy_ablation_ICP} 
reveals that both ICP-MS and XRF yield sub-optimal results for TD. This is in line  with laboratory practice, as elemental analysis methods are not considered meaningful for detecting standard artificial heat treatment of sapphires.

\subsection*{Supplementary Note 2: A closer look at per-class performance}
\begin{figure}[H]
    \centering
    \includegraphics[scale = 0.2]{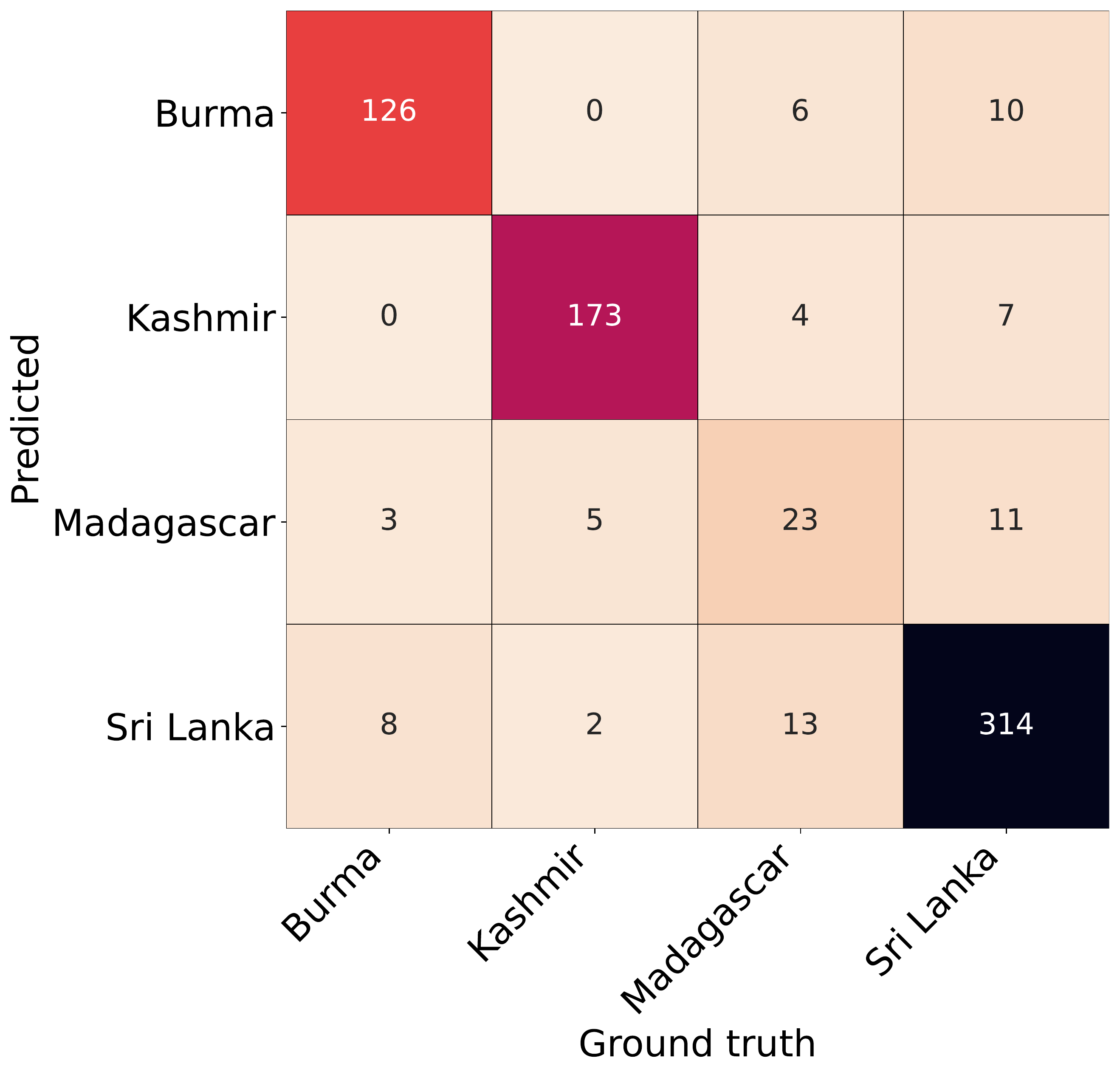}
    \includegraphics[scale = 0.2]{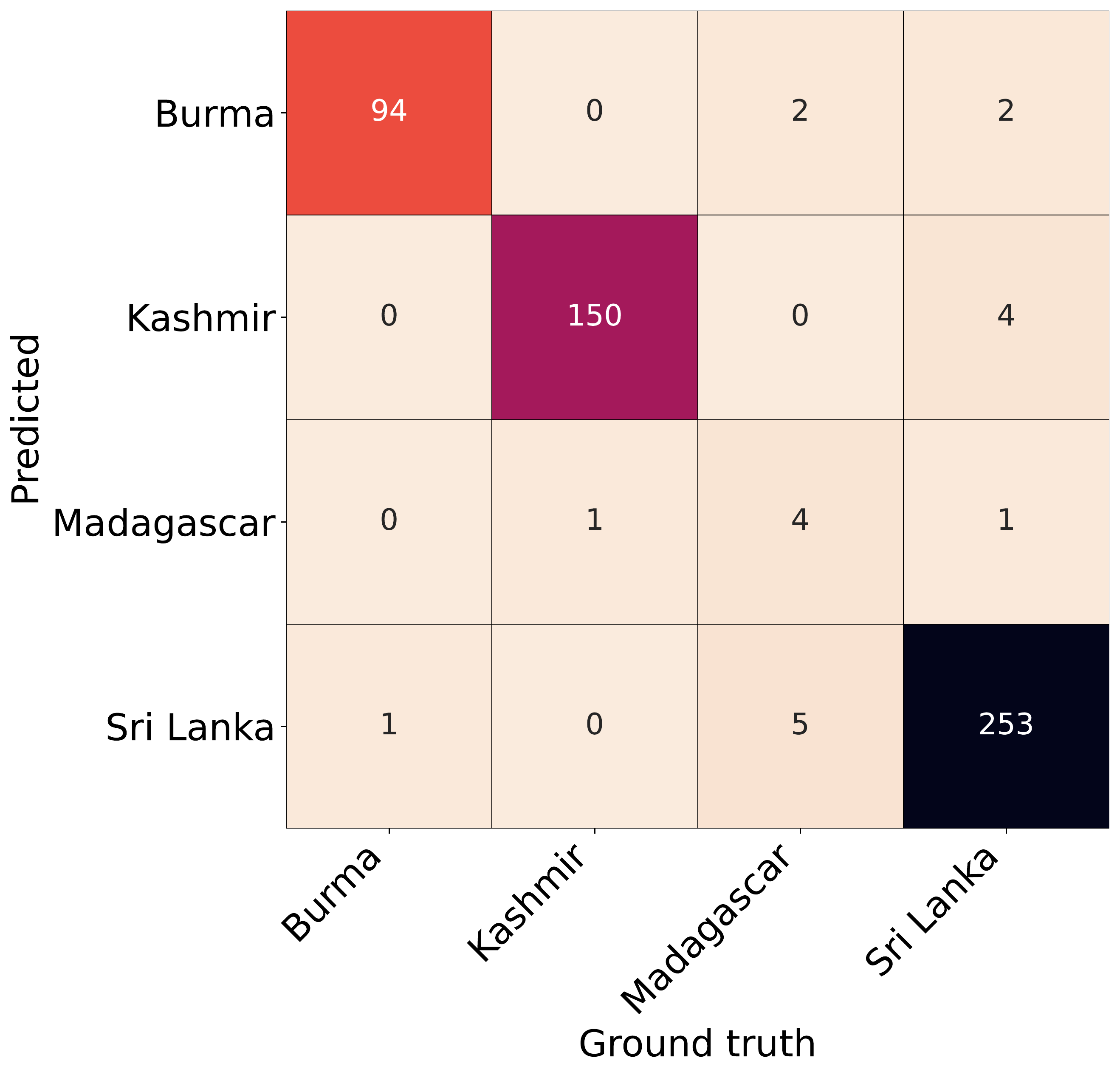} 
    \includegraphics[scale = 0.2]{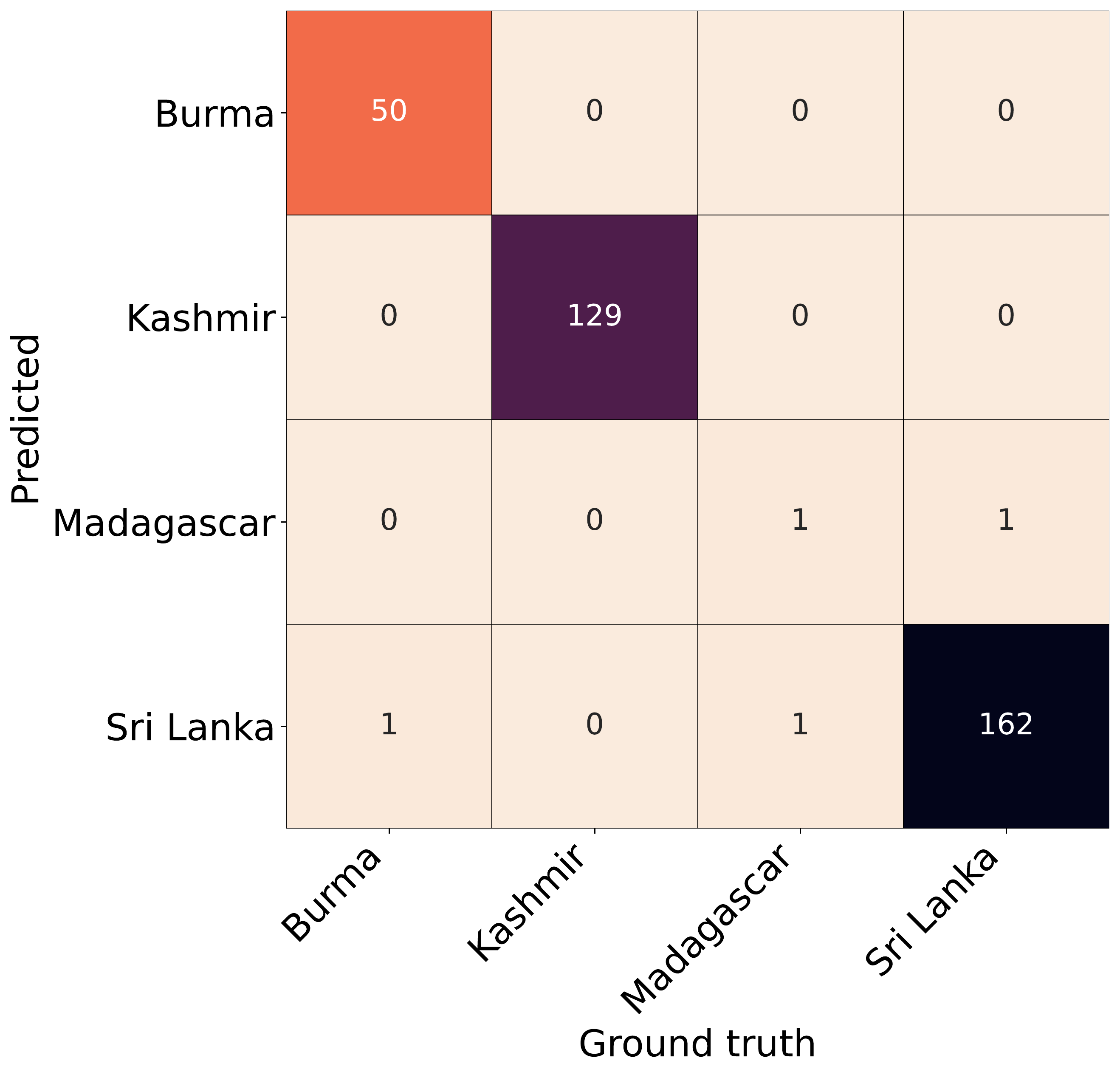} 
    
    \caption{Confusion matrices for OD in the three considered operating modes, namely (Left) \texttt{None}, (Middle) \texttt{Mode 1}, and (Right) \texttt{Mode 2}.
    }  
    
    \label{fig:sapphire_origin_accuracy_confusion}
\end{figure}

\begin{figure}[H]
    \centering
    \includegraphics[scale = 0.2]{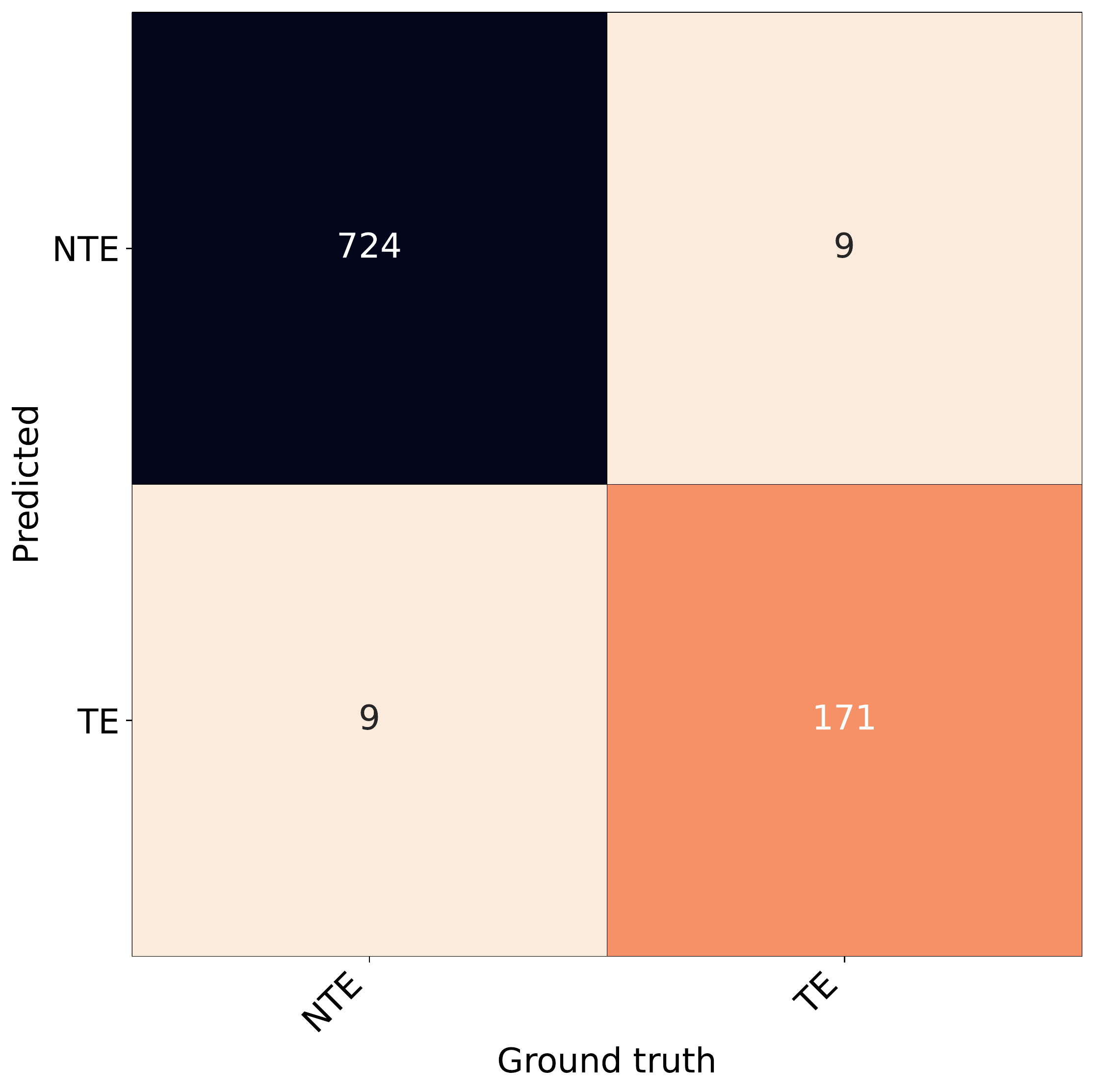} 
    \includegraphics[scale = 0.2]{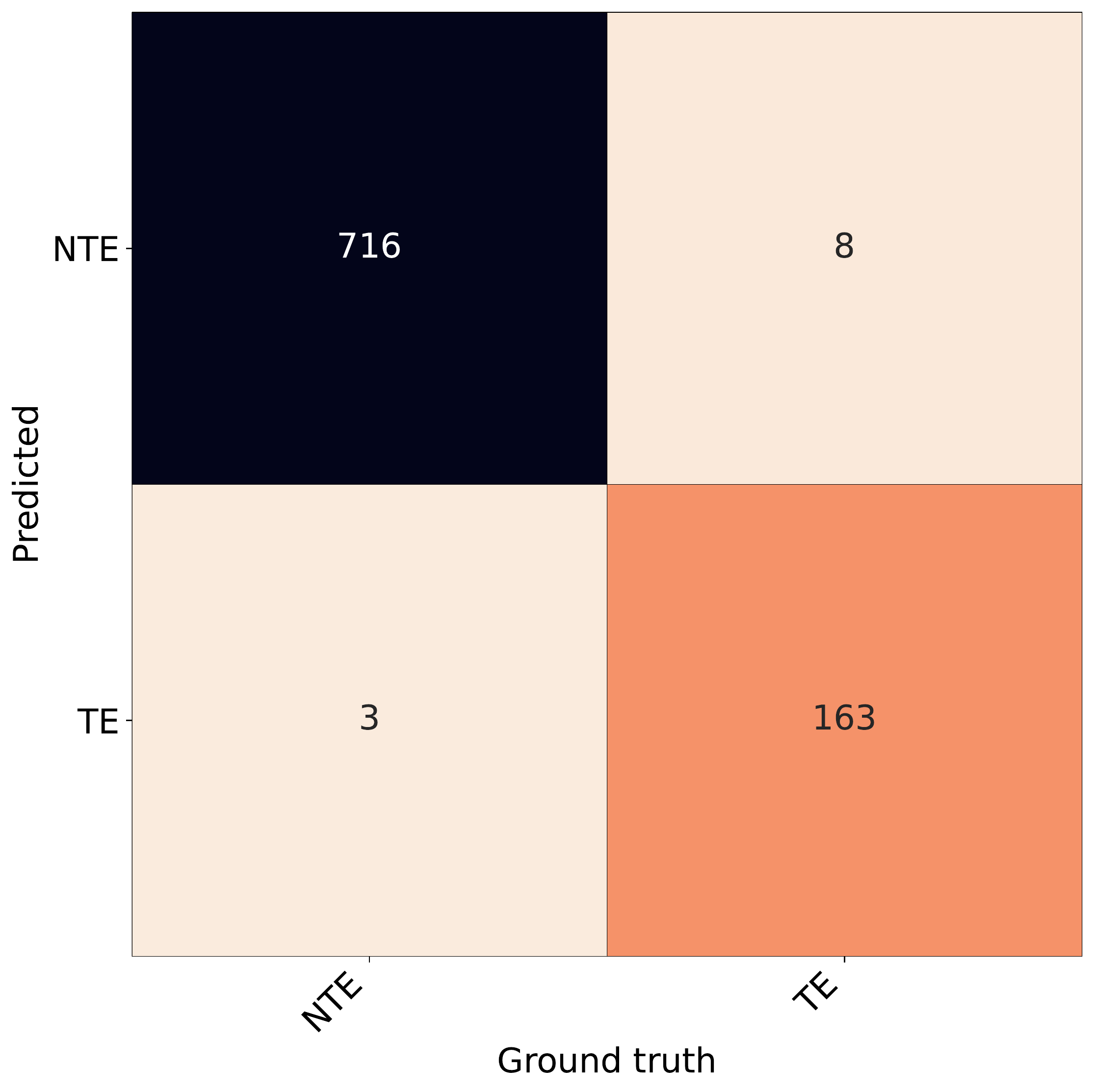} 
    \includegraphics[scale = 0.2]{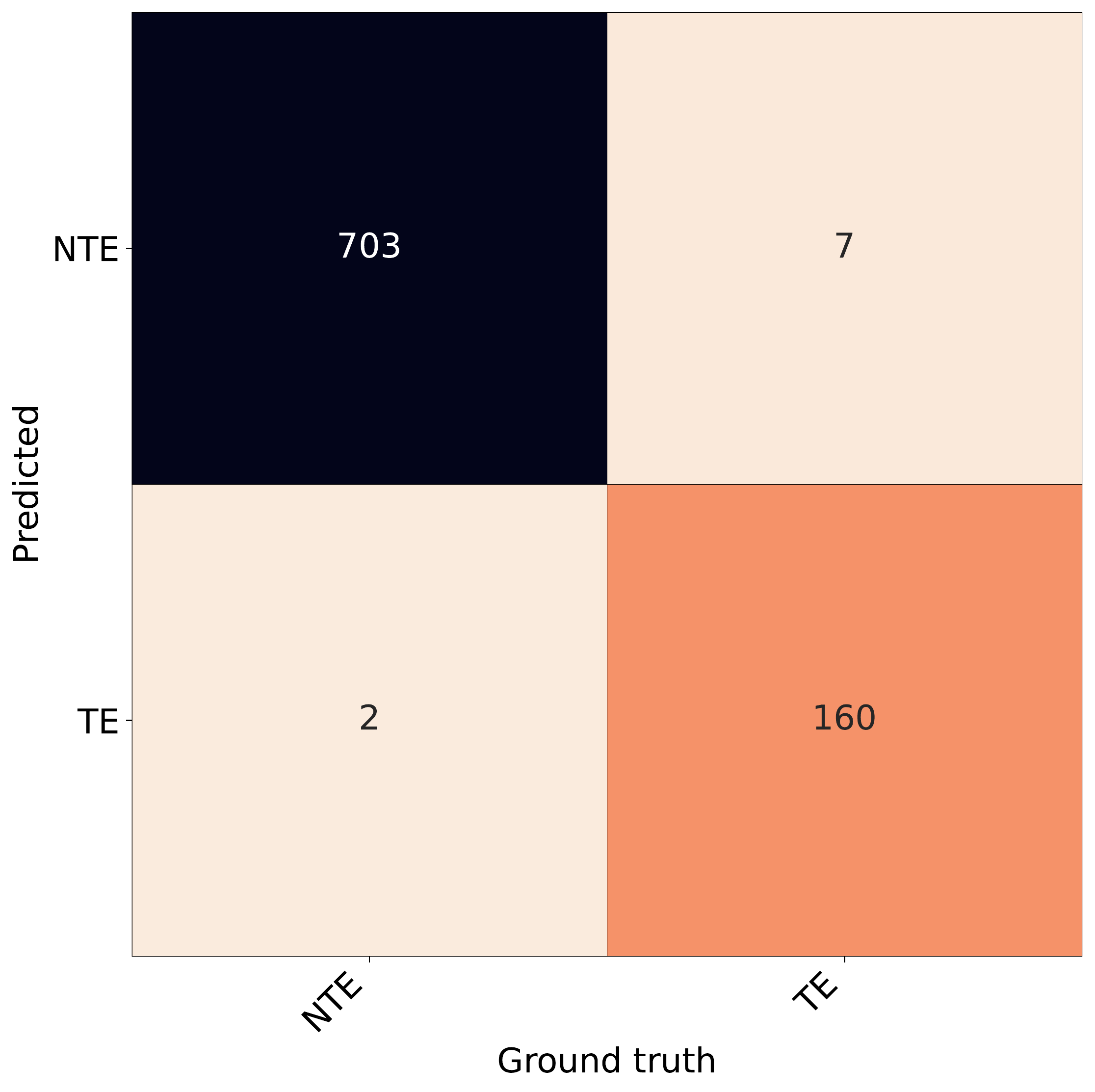} 
    
    \caption{Confusion matrices for TD in the three considered operating modes, namely (Left) \texttt{None}, (Middle) \texttt{Mode 1}, and (Right) \texttt{Mode 2}.
    }  
    
    \label{fig:sapphire_heat_accuracy_confusion}
\end{figure}

In this Note, we present the per-class performance of \gemnet{} for both OD and TD. Supplementary Fig. \ref{fig:sapphire_origin_accuracy_confusion} and \ref{fig:sapphire_heat_accuracy_confusion} display the confusion matrices for OD and TD in three operational modes (\texttt{None}, \texttt{Mode~1}, and \texttt{Mode~2}). The OD confusion matrices reveal that \gemnet{} performs well on stones from Burma, Kashmir, and Sri Lanka while its performance on stones from Madagascar is less accurate. This is expected, as Madagascar stones possess highly diverse features, making consistent OD challenging \cite{krebs2020evaluation}. In the highest confidence mode, \gemnet{} achieves nearly zero errors for stones from Burma, Kashmir, and Sri Lanka, while reducing the number of predictions for Madagascar stones; indicating a lower confidence. The TD confusion matrices of \gemnet{} demonstrate a similar distribution of errors between treated and non-treated stones. Remarkably, by using the high confidence mode, the number of errors can be halved, from 18 in mode \texttt{None} to 9 in \texttt{Mode 2}, while still allowing the processing of a majority of treated and non-treated stones.

\subsection*{Supplementary Note 3: Comparison with masked model}
\begin{figure}[H]
    \centering
    
    \includegraphics[scale = 0.45]{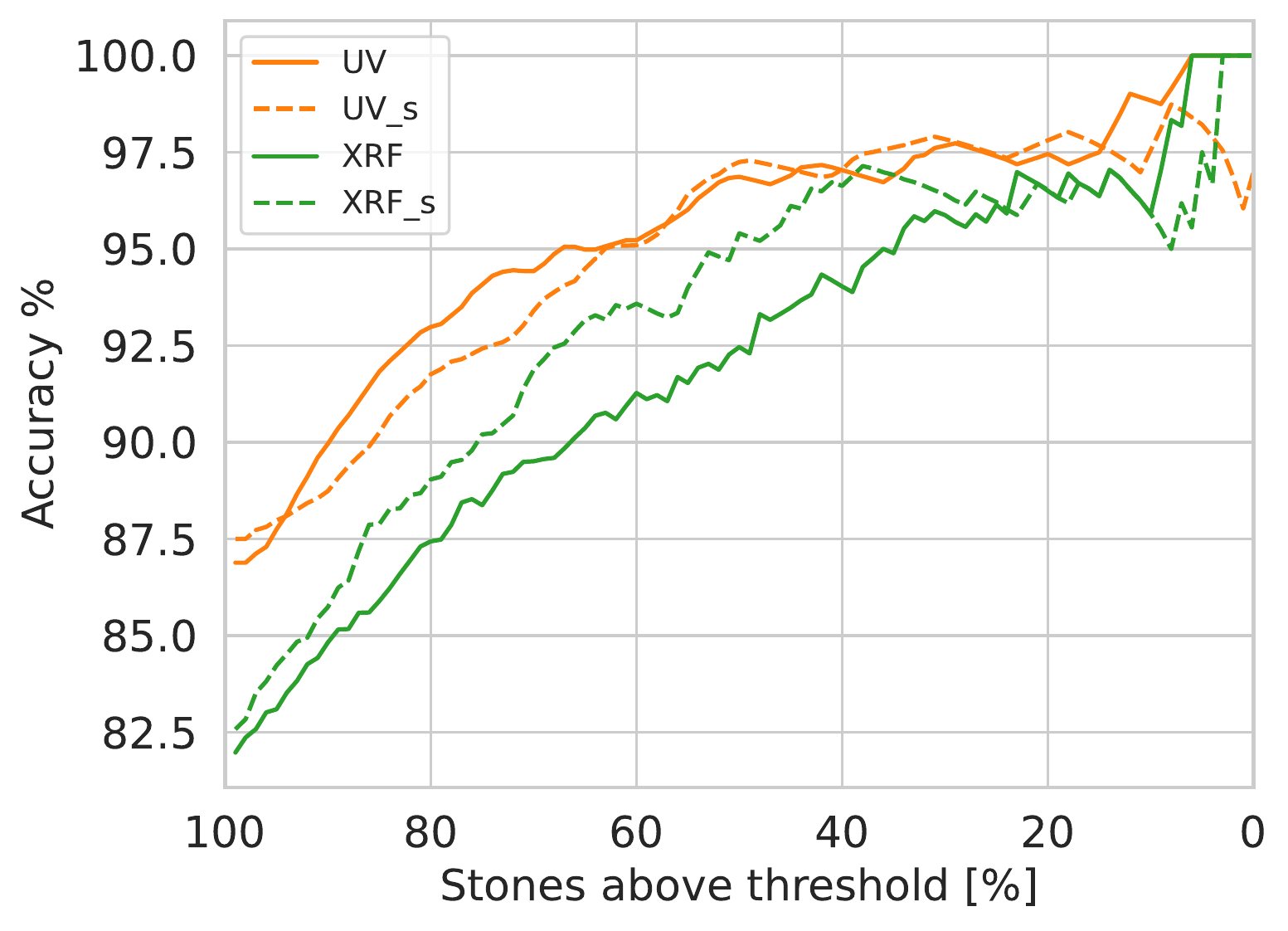} 
    \includegraphics[scale = 0.45]{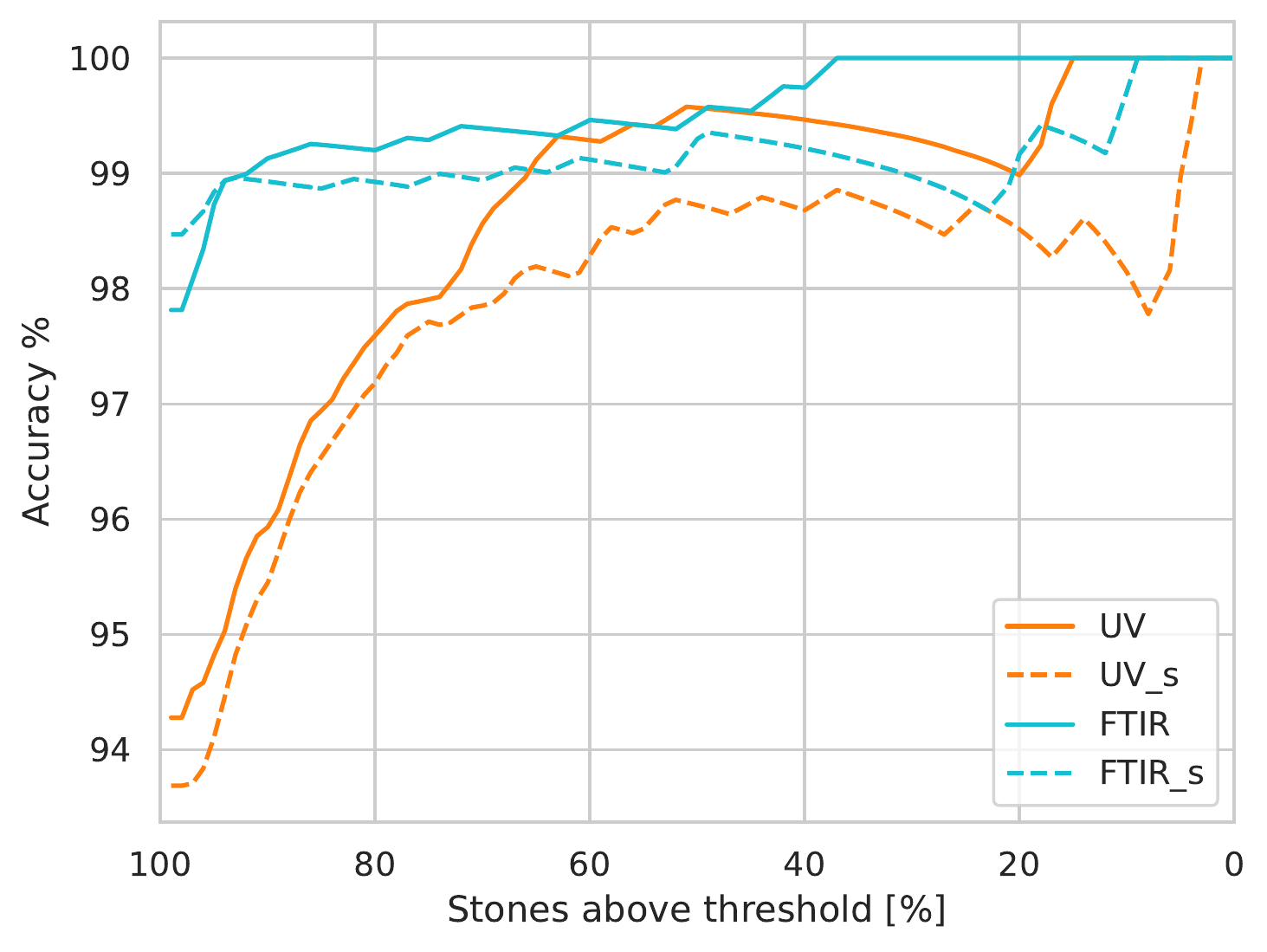} 
    
    \caption{Accuracy [\%] vs.\ stones above the threshold [\%] for (Left) OD and (Right) TD. Dashed lines are used to indicate the models trained exclusively on a single data source while full lines indicate \gemnet{} appropriately masked to hide non-accessible data sources.
    }  
    
    \label{fig:sapphire_accuracy_ablation_masked}
\end{figure}

This Note investigates the impact of employing our masking mechanism (illustrated in Fig.~\ref{fig:main} with switch symbols) when utilizing  \gemnet{} on a subset of all the possible data sources. The masking mechanism involves eliminating a portion of the input data sources.  The primary objective of this study is to compare the effectiveness of the standard \gemnet{} version, which is trained on all data sources and then masked, with that of models trained on a single data source. Supplementary Fig.~\ref{fig:sapphire_accuracy_ablation_masked} shows that the masked model's performance is comparable to that of the models trained on individual data sources. This finding implies  that \gemnet{} can be efficiently applied in situations where only a subset of data sources is available.

\subsection*{Supplementary Note 4: Comparison with an ensemble model}
\begin{figure}[H]
    \centering
    
    \includegraphics[scale = 0.45]{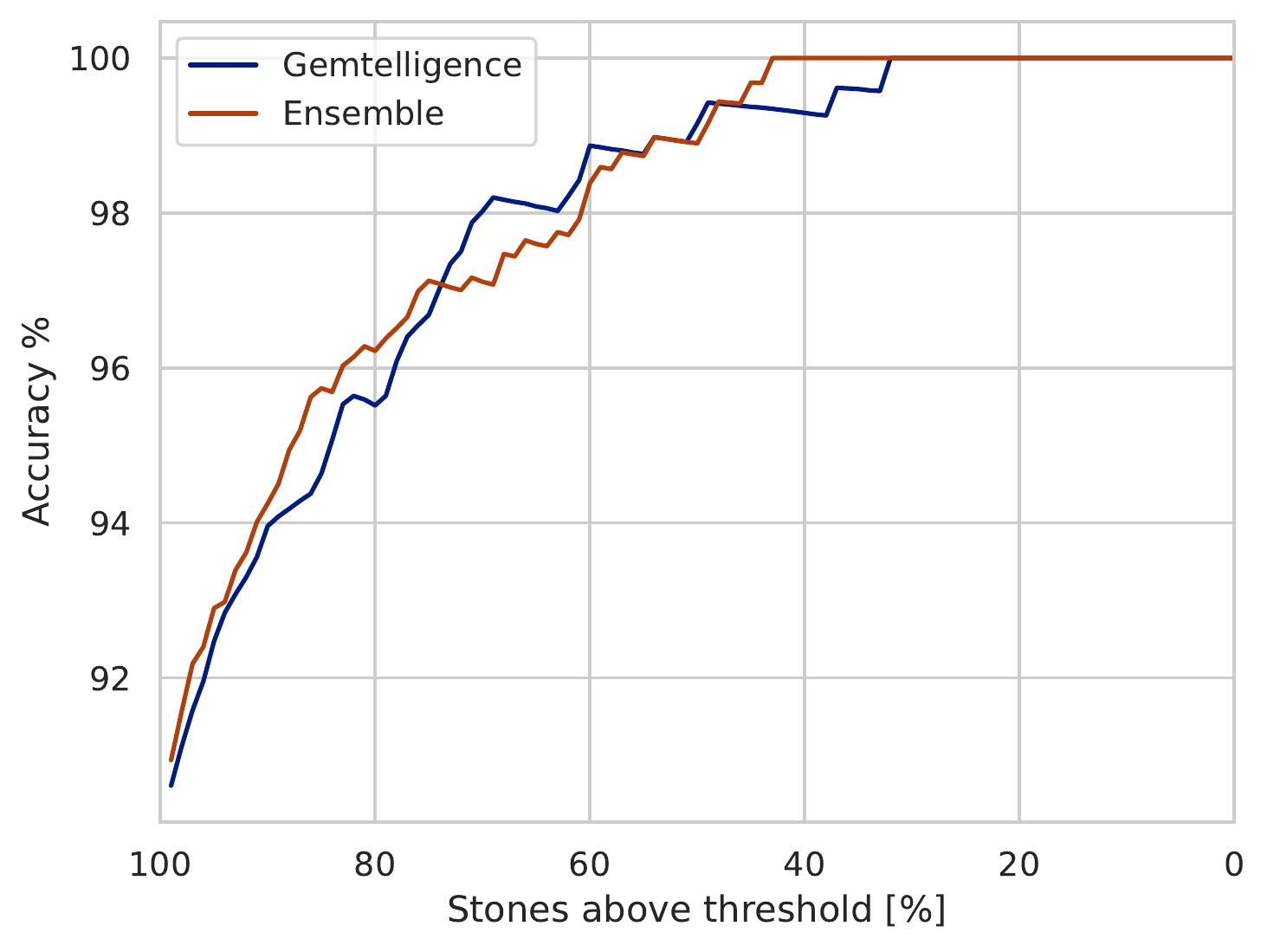} 
    \includegraphics[scale = 0.45]{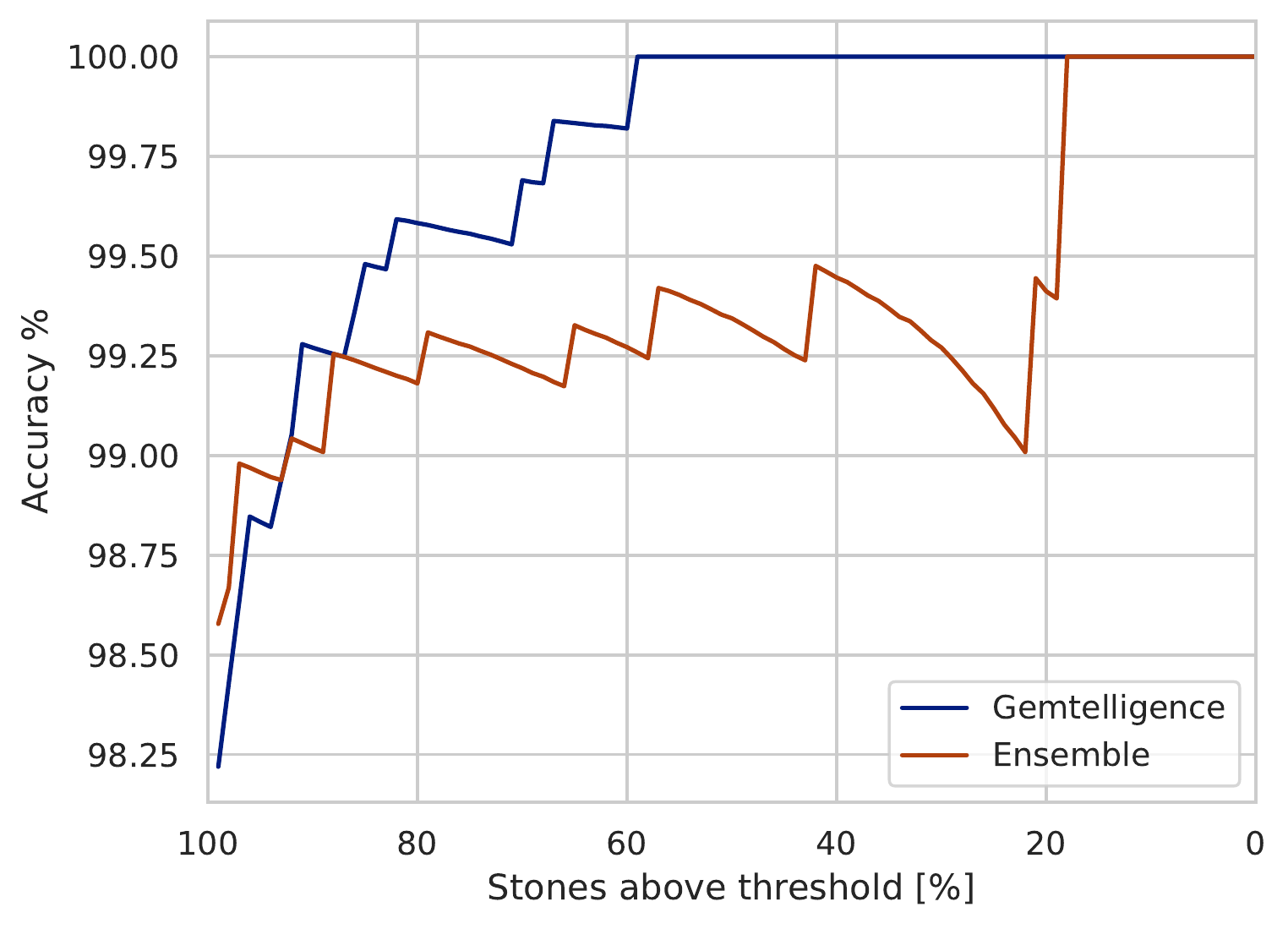} 
    
    \caption{Accuracy [\%] vs.\ stones above the threshold [\%] for (Left) OD and (Right) TD. Both panels show a comparison between the ensemble model (orange) and the \gemnet{} (blue).
    }  
    
    \label{fig:sapphire_accuracy_ablation_ensemble}
\end{figure}

This Note investigates the impact of processing all data sources end-to-end in the \gemnet{} architecture, instead of utilizing a set of classifiers trained separately on each data source. To obtain a single prediction, we train a meta-model consisting of a linear layer followed by a softmax operator, which takes the predictions of the single classifiers' (corresponding to the different encoders used in \gemnet{})  as input and generates the final result. This experiment aims to understand how \gemnet{} combines information from different data modalities and at which network hierarchy level such fusion occurs. Supplementary Fig. \ref{fig:sapphire_accuracy_ablation_ensemble} presents a comparison of the \gemnet{} and ensemble model's performance on the OD and TD tasks. The results show that the two models perform similarly on both tasks, with \gemnet{} achieving slightly better results on TD. This outcome suggests that while the model effectively combines the representations of various input modalities, the encoders mainly extract coarse high-level features, focusing on more abstract  or high-level features. 
This could be attributed to the model being operated in a low data regime (only a few thousand input samples).
Nonetheless, the end-to-end design of \gemnet{} is expected to offer even more significant benefits as the training set grows larger \cite{jaegle2021perceiver}, which is anticipated with upcoming data acquisitions. 
Moreover, using a single end-to-end model offers several practical advantages such as simplicity, maintainability, and robustness, which make it a preferred choice for real-world applications.

\subsection*{Supplementary Note 5: Dataset details and composition}
Between 2013 and 2020, three laboratories located in Lucerne, New York, and Hong Kong, as well as a mobile laboratory at various gemstone fairs analyzed the stones yielding the dataset. Supplementary Table \ref{table1:number_of_stones} displays the number of stones available for each data source for OD and TD. The difference in the number of OD and TD stones is due to the fact that TD is only conducted on blue sapphires, while OD is performed on request. Furthermore, it should be noted that only the Lucerne laboratory collects ICP-MS data. Also, because of the variation in UV machines used by different labs, the UV data are limited to those obtained in Lucerne, resulting in fewer available data samples for both UV and ICP-MS than other sources.\\
As indicated in Supplementary Table~\ref{table1:number_of_stones}, the test stones constitute only a small number of stones available. This is because the test set is designed to include only stones with available UV, ICP-MS, and XRF data for OD, and UV and FTIR data for TD. Furthermore, for OD, microscopy and ICP-MS sub-conclusions must match, and data are restricted from 2017 onwards. This is because  a standardized method for combining the various measurements (sub-conclusions) was introduced in 2017. Under this new protocol, each data source was analyzed independently, and the resulting sub-conclusions were combined via a weighted mean. This new protocol allowed for a more objective overall evaluation. Before this, the final conclusion was reached in a less standardized way, allowing the gemologist to assign individual importance to each data source, resulting in a final outcome that was susceptible to personal bias regarding the relevance of each data source.

\section*{Supplementary Tables}

\begin{table}[H]
\makebox[0.5\textwidth][c]{%
\begin{tabular}{@{}l|ccc@{}}
                   & \textbf{OD}\qquad\qquad&\qquad\quad\textbf{TD} \\    \hline\hline
\textbf{Available stones}\quad \hspace{-0.3cm} \textit{UV}  &2978 \quad\quad&\qquad\quad 2953 \\ 
\quad\qquad\quad\qquad\quad\textit{FTIR}  &3355 \quad\quad&\qquad\quad 3981  \\ 
\quad\qquad\quad\qquad\quad\textit{XRF}  &4752 \quad\quad&\qquad\quad 5152  \\ 
\quad\qquad\quad\qquad\quad\textit{ICP-MS}  &2809 \quad\quad&\qquad\quad 3071  \\ 
\quad\qquad\quad\qquad\quad\textit{All}  &1419 \quad\quad&\qquad\quad 1452  \\
\quad\qquad\quad\qquad\quad\textit{Any}  &5515 \quad\quad&\qquad\quad 5889  \\  \hline
\textbf{Test stones}\quad\quad\quad \hspace{-0.3cm} \textit{UV}  &705 \quad\quad&\qquad\quad 913 \\ 
\quad\qquad\quad\qquad\quad\textit{FTIR}  &705 \quad\quad&\qquad\quad 913  \\ 
\quad\qquad\quad\qquad\quad\textit{XRF}  &705 \quad\quad&\qquad\quad 913  \\ 
\quad\qquad\quad\qquad\quad\textit{ICP-MS}  &705 \quad\quad&\qquad\quad 913  \\ 
\quad\qquad\quad\qquad\quad\textit{All}  &705 \quad\quad&\qquad\quad 913  \\
\quad\qquad\quad\qquad\quad\textit{Any}  &705 \quad\quad&\qquad\quad 913  \\  \hline

\caption{
Breakdown of the total stones available and the testing subset for task and data source. The term \emph{All} indicates stones with measurements from all data sources (intersection of all the data sources), while \emph{Any} includes stones with measurements from at least one data source (union of the data sources).
}
\label{table1:number_of_stones}
\end{tabular}
}
\end{table}

\section*{Supplementary Figures}

\end{document}